\documentclass[11pt]{article}

\usepackage[preprint]{acl}

\usepackage{times}
\usepackage{latexsym}

\usepackage[T1]{fontenc}

\usepackage[utf8]{inputenc}

\usepackage{microtype}

\usepackage{inconsolata}

\usepackage{graphicx}

\usepackage{multirow}
\usepackage{changes}
\usepackage{tabularx, booktabs, kotex}
\usepackage{amsmath}

\title{K-FinHallu: A Hallucination Detection Benchmark\\for Multi-Turn RAG in Korean Finance}

\author{
  Eunbyeol Cho\thanks{\hspace{1mm}Equal contribution.}$^{1}$,
  Yunseung Lee\footnotemark[1]$^{1,2}$,
  Mirae Kim$^2$ \\
  \textbf{Jeewon Yang}$^1$,
  \textbf{Youngjun Kwak}$^2$,
  \textbf{Edward Choi}$^1$ \\
  $^1$KAIST AI, $^2$Financial Tech Lab, KakaoBank Corp. \\
  \texttt{\{eunbyeol.cho, jwy24, edwardchoi\}@kaist.ac.kr} \\
  \texttt{\{yun.lee, melissa.kim, vivaan.yjkwak\}@lab.kakaobank.com}
}

\begin{document}
\maketitle
\begin{abstract}
Large Language Models (LLMs) have advanced financial automation through Retrieval-Augmented Generation (RAG), yet hallucinations remain a critical barrier to deployment in high-stakes environments. Existing benchmarks focus on single-turn, English-centric tasks, leaving the multi-turn dynamics and linguistic-regulatory nuances of the Korean financial domain unaddressed. We introduce K-FinHallu, the first benchmark for hallucination detection in multi-turn Korean financial RAG. We construct multi-turn dialogues from authentic Korean financial documents and inject hallucinations under a proposed hierarchical taxonomy based on context answerability that explicitly accounts for justified abstention. Benchmarking frontier and open-source LLMs as hallucination detectors, we find that even the strongest models struggle with fine-grained financial diagnostics and refusal behavior. While fine-tuning an 8B model on our training split yields performance competitive with frontier LLMs, justified abstention remains the weakest axis across all evaluated models.
\end{abstract}

\begin{figure}[t!]
  \includegraphics[width=\columnwidth]{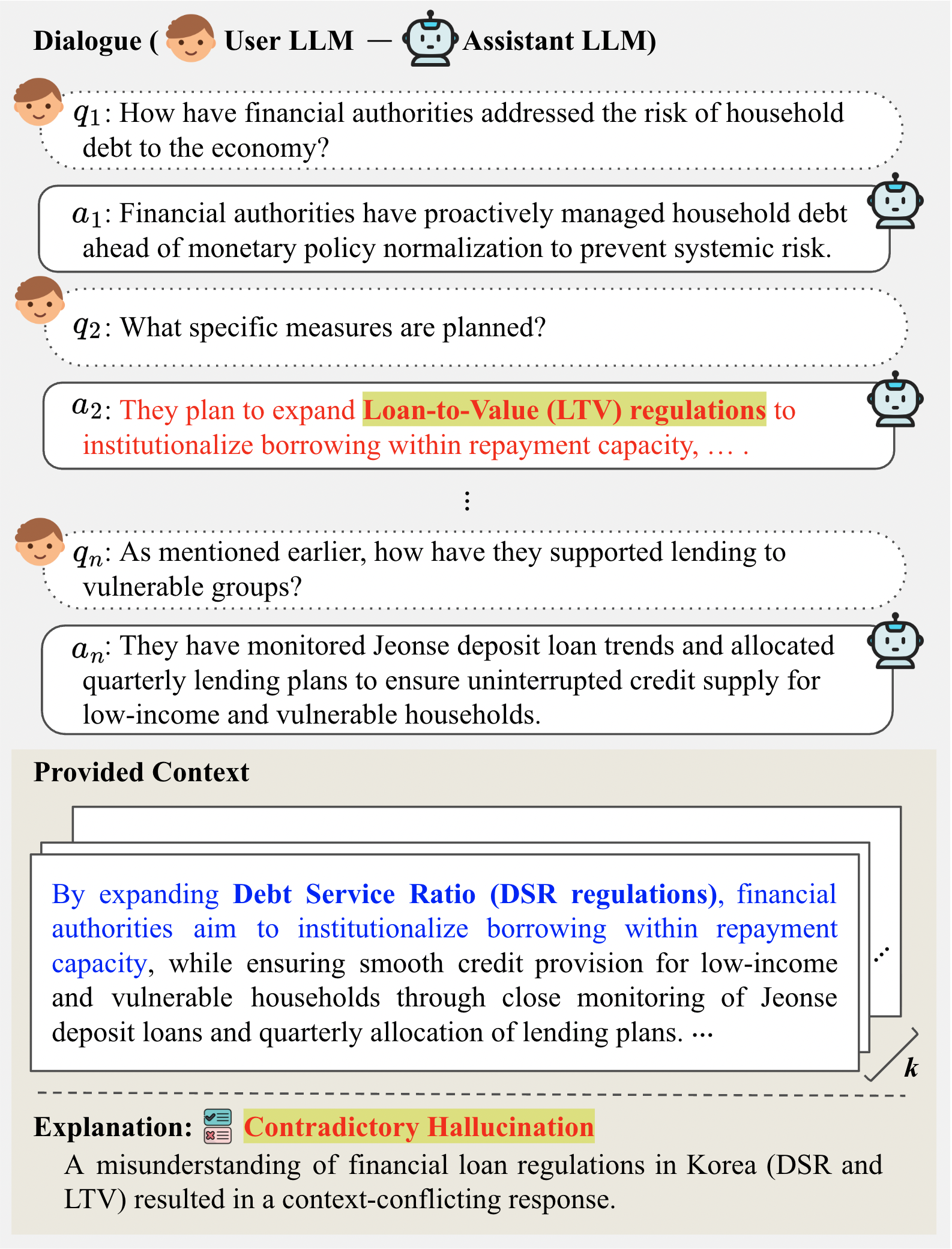}
  \caption{Example of a contradictory hallucination in K-FinHallu. Hallucinated content is shown in red, hallucination spans are highlighted, and blue text indicates the clue text.}
  \label{fig:dataset_overview}
\end{figure}

\section{Introduction}

The rapid advancement of Large Language Models (LLMs)~\cite{brown2020language} has catalyzed a paradigm shift across various industries. In the financial sector, LLMs are increasingly deployed to enhance operational efficiency in areas such as market analysis and the summarization of complex financial reports.
However, the financial domain demands a level of precision and real-time information that often exceeds an LLM's internal parametric knowledge.
Consequently, Retrieval-Augmented Generation (RAG) has emerged as the standard architecture for financial AI~\cite{lithgow2025assessing}, providing evidence-based responses grounded in external, reliable knowledge bases.

Despite these advantages, LLMs still produce \textit{hallucinations}— outputs that contradict retrieved documents or lack contextual grounding~\cite{shuster2021retrieval, niu2024ragtruth}. In finance, such hallucinations are particularly critical as even small factual errors can lead to substantial monetary losses, compliance risks, and the erosion of user trust. Therefore, establishing rigorous evaluation benchmarks to detect domain-specific hallucinations is essential for reliable deployment.

While existing works have addressed hallucinations in the context of RAG, two primary gaps remain in reflecting real-world financial service environments. \textbf{First, there is a lack of established evaluation frameworks for multi-turn conversational RAG within the financial domain.}
Financial services typically operate through continuous interactions (\textit{e.g.}, chatbots) involving anaphora and implicit references.
Existing benchmarks, which focus predominantly on single-turn tasks~\cite{niu2024ragtruth,mishra2024fine}, struggle to identify hallucinations arising from context-tracking failures, such as logical inconsistencies with prior turns or the misinterpretation of implicit references.
\textbf{Second, the unique nature of the Korean financial environment is underrepresented in existing benchmarks.}
Existing benchmarks~\cite{phantom, wang2025omnieval} focus on Western institutional contexts. These datasets fail to account for Korea-specific financial systems, such as \textit{Jeonse} lease system~\cite{Yun2020Jeonse}, or local regulatory environments.
Furthermore, Korean linguistic features—including its agglutinative morphology and the polysemy of Sino-Korean technical terminology—cannot be addressed through simple translation.
To the best of our knowledge, there is currently no native Korean financial benchmark specifically designed to evaluate hallucinations in multi-turn dialogues.

To bridge these gaps, we introduce \textbf{K-FinHallu}, a multi-turn RAG hallucination detection benchmark for the Korean financial domain, constructed from real-world Korean financial documents.
We build K-FinHallu through a pipeline of faithful dialogue generation and systematic hallucination injection, covering numerical perturbations, cross-turn inconsistencies, and failures to abstain when retrieval is unsuccessful.
Figure~\ref{fig:dataset_overview} illustrates a representative example of a contradictory hallucination from K-FinHallu, where the model misinterprets specific financial loan regulations.
Benchmarking frontier and open-source LLMs as detectors, we find that even the strongest models struggle with fine-grained financial diagnostics and refusal behavior. Furthermore, we demonstrate that our pipeline can scale to automatically construct a training split, on which a fine-tuned 8B model with rationale supervision matches or outperforms frontier LLMs.
Our primary contributions are as follows:
\begin{itemize}\itemsep0pt \topsep2pt 
    \item We release \textbf{K-FinHallu}, the first multi-turn hallucination detection benchmark for Korean financial RAG, built on authentic Korean financial documents.
    \item We propose a \textbf{hierarchical hallucination taxonomy} based on context answerability that explicitly accounts for justified abstention in multi-turn RAG dialogues.
    \item We demonstrate that our construction pipeline can automatically generate a training split, on which a fine-tuned 8B model with rationale supervision matches or outperforms frontier LLMs.
\end{itemize}
\vspace{-2pt}

\begin{figure*}[t]
  \includegraphics[width=0.98\linewidth]{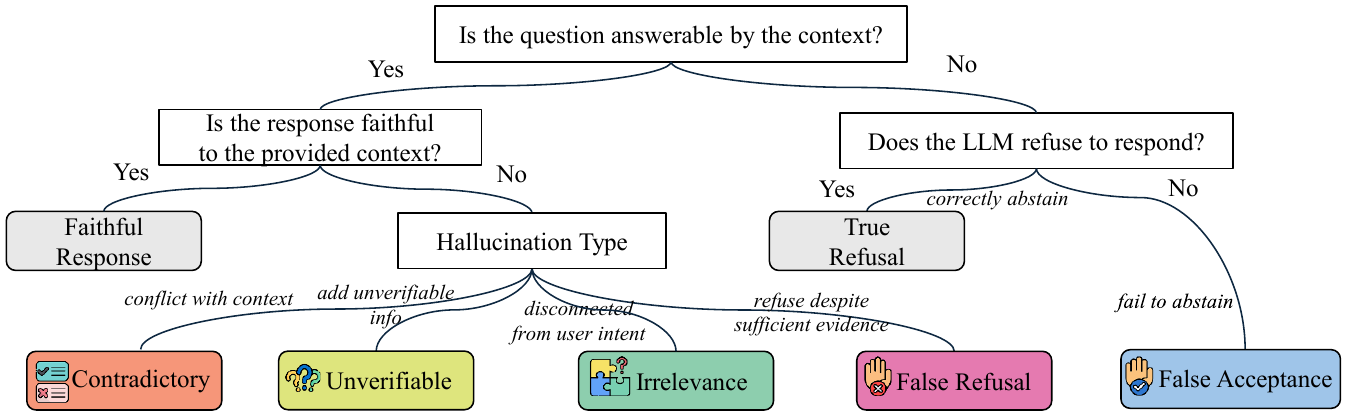}
  \caption {Hallucination taxonomy for multi-turn RAG. We categorize five types of hallucinations based on context answerability: four types of groundedness failures in answerable scenarios where sufficient evidence is provided, and one type of fabrication in unanswerable scenarios.}
  \label{fig:taxonomy}
\end{figure*}

\section{Related Work}

\subsection{Hallucination Detection Benchmarks}
Recent studies have actively explored hallucination evaluation in LLMs~\cite{mishra2024fine,seo2025k}. HaluEval \cite{li2023halueval} introduced a foundational benchmark for tasks such as question answering and summarization, but it primarily focuses on single-turn interactions and short contexts.
RAGTruth \cite{niu2024ragtruth} addresses hallucinations in RAG systems by detecting them across various domains. However, it focuses on the accuracy of individual responses rather than the flow of multi-turn dialogues.
For dialogue-level evaluations, DiaHalu \cite{chen2024diahalu} and FaithDial \cite{dziri2022faithdial} assess the consistency and factuality of general conversations. HalluLens \cite{bang2025hallulens} emphasizes the distinction between factuality and hallucination, while also noting the lack of evaluation for cases where a model should refuse to answer. 
Our research builds on these findings by focusing on intrinsic hallucination rather than external knowledge. We specifically evaluate the logical consistency with retrieved documents and the ability to abstain from answering in multi-turn dialogues.

\subsection{Financial Hallucination Benchmarks}
The financial domain requires high precision and up-to-date information, making specialized benchmarks essential.
Existing benchmarks such as AlphaFin \cite{li2024alphafin} and FinBen \cite{xie2024finben} focus on measuring financial analysis and information retrieval performance. However, they do not typically verify the presence of hallucinations in generated responses.
While OmniEval \cite{wang2025omnieval} evaluates RAG in finance through multi-dimensional metrics using synthetic data, including hallucinations, it tends to overlook the context degradation inherent in long-form financial reports.
PHANTOM \cite{phantom} introduced a benchmark for hallucination detection in long-context financial QA by focusing on source faithfulness, which measures whether a response is strictly based on the provided documents.
However, this benchmark is primarily based on Western financial systems and English datasets. Although financial services are often provided through chatbots, existing benchmarks do not sufficiently reflect hallucinations caused by context loss or anaphora resolution in multi-turn dialogues.

\subsection{Korean Hallucination Benchmarks}
Benchmarks for evaluating Korean LLMs have progressed through datasets such as KMMLU \cite{son2025kmmlu} and HAE-RAE \cite{son2024hae}. Despite this progress, datasets specifically designed for hallucination evaluation remain limited. Existing benchmarks such as Ko-TruthfulQA \cite{lin2022truthfulqa, park2024open} often rely on translated English datasets, which may fail to capture Korean linguistic and cultural nuances~\cite{hendrycks2020measuring, seo2024kocommongen}.
K-HALU \cite{seo2025k} addresses this by covering domains such as news and books within a Korean context.
However, it does not account for financial regimes unique to Korea, such as housing finance regulations. Our work is the first framework to consider these regional characteristics and linguistic complexities while evaluating multi-turn hallucinations.

\section{Task Definition}

\subsection{Task Formulation}
\label{subsec:taskformulation}
The primary objective of K-FinHallu is to detect hallucinations in multi-turn RAG dialogues. Formally, we define a multi-turn RAG dialogue at turn $n$ as a tuple $(H_n, q_n, \mathcal{P}_n)$, where $H_n = ((q_1, a_1), \dots, (q_{n-1}, a_{n-1}))$ is the ordered dialogue history, $q_n$ is the current user query, and $\mathcal{P}_n = \{p_n^1, p_n^2, \dots, p_n^K\}$ is a set of $K$ retrieved passages. The model $f$ generates a response $a_n = f(H_n, q_n, \mathcal{P}_n)$ that should be strictly grounded in the provided context $H_n \cup \mathcal{P}_n$. The task is to predict whether $a_n$ is faithful or hallucinated at turn $n$, given $(H_n, q_n, \mathcal{P}_n, a_n)$.

To detect hallucinations independently of the retrieval system's performance, we adopt a \textit{simulation-based RAG} framework. Instead of employing a live retrieval module which may introduce uncontrolled retrieval errors, the model is provided with a controlled set of documents $\mathcal{P}_n$. This setup enables systematic manipulation of evidence presence, such as intentionally substituting the relevant passage with a \textit{hard negative} passage that is semantically similar to the query $q_n$ but lacks the critical information required for an accurate response.

In this work, we focus exclusively on detecting \textit{intrinsic hallucinations} based on source faithfulness, in line with~\cite{phantom}. This is particularly vital in the financial domain, where information is time-sensitive and conclusions must be derived from authoritative evidence rather than the model's parametric knowledge. We assume that all queries in our benchmark require RAG, involving specialized expertise that cannot be addressed without the external context.

\subsection{Hallucination Taxonomy}
\label{sec:taxonomy}
We establish a hierarchical taxonomy of hallucinations in multi-turn RAG dialogues, illustrated in Figure~\ref{fig:taxonomy}.
The top-level criterion is \textit{answerability}, defined as whether the context $H_n \cup \mathcal{P}_n$ contains sufficient evidence to address $q_n$.
When the query is \textit{answerable}, hallucinations are categorized into four types:

\begin{itemize}

    \item \textbf{Contradictory}:
     The response conflicts with the provided context, which includes both the dialogue history $H_n$ and the retrieved passages $\mathcal{P}_n$.

    \item \textbf{Unverifiable}: The response includes external claims or general knowledge that is not supported by the given $\mathcal{P}_n$.

    \item \textbf{Irrelevance}: The response is logically disconnected from the user's intent or $q_n$, even if the facts themselves are present in the document.

    \item \textbf{False Refusal}: The model refuses to answer despite sufficient evidence being present in $\mathcal{P}_n$.

\end{itemize}

When a query is \textit{unanswerable} due to a lack of evidence in $\mathcal{P}_n$, hallucination occurs if the model fails to withhold its response, defined as \textbf{False Acceptance}. Conversely, \textbf{True Refusal} represents the correct behavior. Together, they measure the model's capacity for honest abstention—judging when to provide an answer versus when to acknowledge missing evidence.

Empirically, we observe all five types in both naturally generated outputs from frontier LLMs and failure cases from a deployed Korean financial RAG service, detailed in Appendix~\ref{appendix:realism}.

\begin{figure*}[t]
  \includegraphics[width=0.98\linewidth]{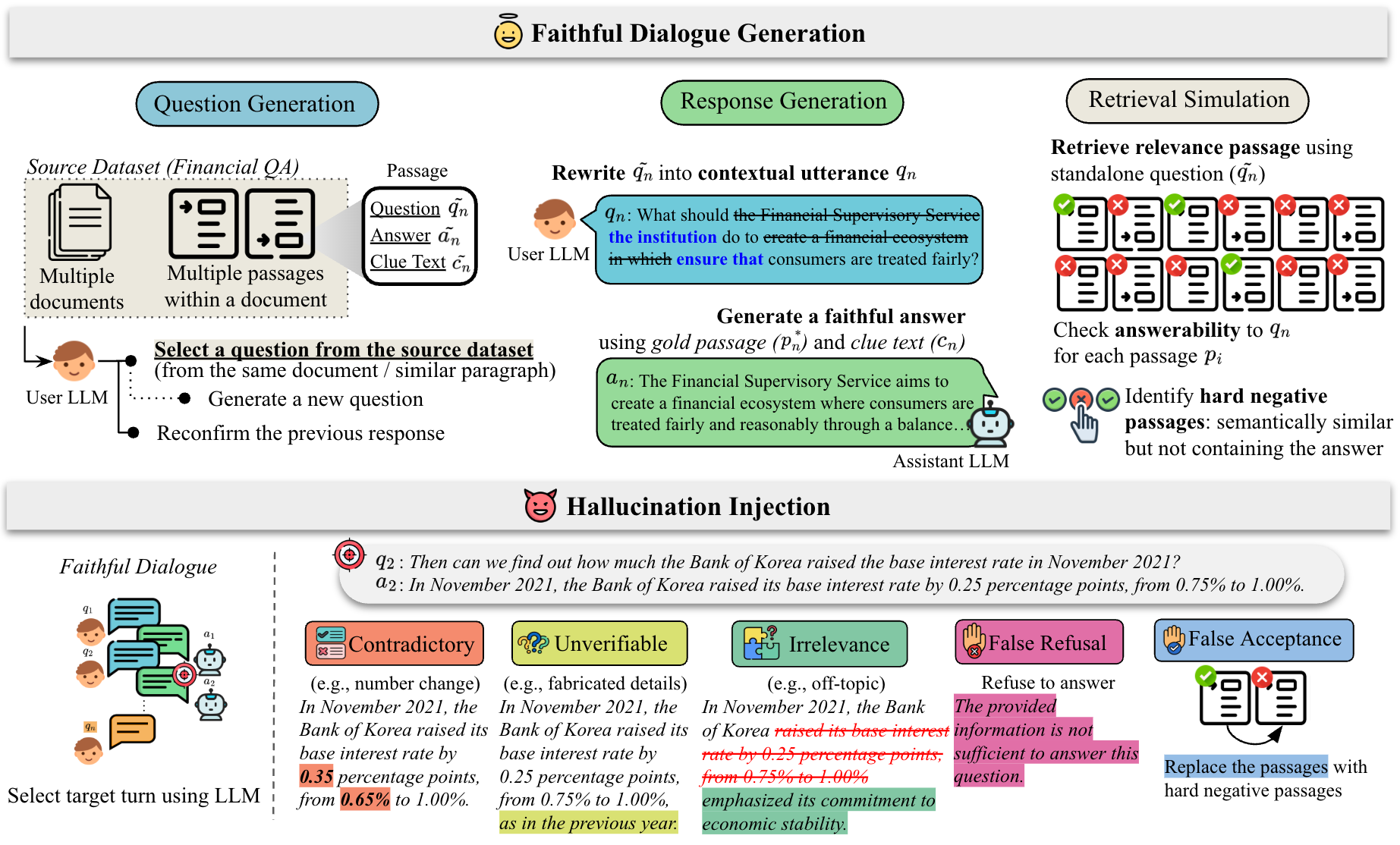}
  \caption {Overview of the K-FinHallu construction pipeline. We first generate faithful dialogues and then inject hallucinations to construct the final dataset. Highlighted text illustrates how faithful responses are transformed under different hallucination types.}
  \label{fig:dataset_generation_overview}
\end{figure*}

\section{K-FinHallu}

\subsection{Source Dataset}
We use the publicly available ``Korean Financial and Legal Document Machine Reading Comprehension'' dataset from AI-Hub as the source corpus, filtered to its financial category.\footnote{AI-Hub is a public data repository operated by Korea's National Information Society Agency (NIA): \url{https://aihub.or.kr/aihubdata/data/view.do?dataSetSn=71610}}
Each source document contains multiple passages $p$, each associated with a set of triplets $\{(\tilde{q}^{(i)}, \tilde{a}^{(i)}, \tilde{c}^{(i)})\}_{i=1}^m$, where $\tilde{q}^{(i)}$ is a question, $\tilde{a}^{(i)}$ is the reference answer, and $\tilde{c}^{(i)}$ is the \textit{clue text}—the evidentiary span within $p$ required to derive $\tilde{a}^{(i)}$.
Passages are segmented to a maximum of 1{,}000 tokens using the KURE Korean embedding model~\cite{KURE}.
To prevent data leakage, the corpus is partitioned by source institution: the test corpus is drawn from the Financial Supervisory Service (FSS) and the Korea Consumer Agency, while the training corpus is drawn from disjoint institutions (Appendix~\ref{appendix:dataset_distribution}). The test and training splits contain 272 / 2{,}732 documents and 2{,}064 / 42{,}364 queries, respectively.

\subsection{Faithful Dialogue Generation}

We design a dialogue simulation pipeline involving a user LLM and an assistant LLM (both powered by GPT-4o) to construct \textit{faithful dialogues}—hallucination-free conversations that serve as the basis for hallucination injection. Figure~\ref{fig:dataset_generation_overview} shows an overview of the pipeline.
Each dialogue is initiated with a seed question sampled from the source dataset.

\paragraph{Query Generation}
The user LLM selects a candidate question $\tilde{q}_n$ from existing questions in the same document or in the top-10 most similar passages by embedding similarity, maintaining a coherent conversational flow.
If no relevant candidate exists, the user LLM generates a new question based on the current passage.
In addition, to evaluate context tracking across turns, we introduce a \textit{reconfirm} question after the dialogue has sufficiently progressed ($n \geq 4$). These turns are randomly inserted to prompt the assistant to verify previous statements in $H_n$.
Finally, the selected $\tilde{q}_n$ is rewritten into a natural, context-dependent utterance $q_n$ by incorporating stylistic paraphrasing, coreference, and ellipsis.

\paragraph{Response Generation}
The assistant LLM generates each response $a_n$ grounded in the gold passage $p_n^*$ and clue text $c_n$ from the source dataset, following the approach of CoQA~\cite{reddy2019coqa}. This is intended to yield a faithful dialogue that serves as a clean substrate for hallucination injection.

\paragraph{Retrieval Simulation}
For each turn, we construct a retrieval context $\mathcal{P}_n$ to simulate realistic RAG settings. To bypass the ambiguity of conversational queries $q_n$ (e.g., anaphora), we use the decontextualized standalone $\tilde{q}_n$ as the retrieval key.
In addition, we refresh $\mathcal{P}_n$ with only the top-$K$ ($K{=}10$) passages for the current turn, rather than accumulating across turns, to prevent retrieval noise and maintain information density~\cite{katsis2025mtrag}.

We vary the composition of $\mathcal{P}_n$ by answerability, following Section~\ref{sec:taxonomy}.
For \textit{answerable} scenarios, $\mathcal{P}_n$ contains the gold passage $p^*$ (and possibly other positive passages $p^+$ that also support the query). For \textit{unanswerable} scenarios, $\mathcal{P}_n$ consists entirely of hard negatives ($p^-$)—passages semantically similar to $\tilde{q}_n$ that lack the evidence required for a response.
Hard negatives are selected by a two-step automated pipeline that retains passages where the answer cannot be derived from the context and the clue text is absent (Appendix~\ref{appendix:passage_classifier}), followed by manual verification.

\paragraph{Quality Control}
We evaluate the generated dialogues using an LLM-as-a-judge across five criteria: \textit{answerability}, \textit{plausibility}, \textit{correctness}, \textit{diversity}, and \textit{coherency}~\cite{lee2024multi}. Dialogues scoring below a predefined threshold on any criterion are filtered out. We validate the reliability of this LLM judge against four human annotators and observe robust agreement based on Gwet's AC1\footnote{We use Gwet's AC1 over Cohen's $\kappa$ to avoid the kappa paradox under high agreement with skewed class distributions~\cite{gwet2008computing}.} (Table~\ref{tab:faithful_dialogue_agreement}). This filtering retains 74.3\% of test and 77.0\% of train dialogues. For the test set, human annotators further review all retained dialogues and revise as needed (details in Appendix~\ref{appendix:annotation_ui}).

\subsection{Hallucination Injection}
\label{sec:injection}

To construct a challenging benchmark, we develop an automated pipeline to inject subtle yet critical hallucinations into faithful dialogues. Our injection strategy is designed with financial-domain experts and informed by real-world hallucination patterns observed in deployed Korean financial RAG systems (Appendix~\ref{appendix:realism_fgt}).

The pipeline identifies contextually optimal turns for injection based on hallucination type (e.g., \textit{Contradictory} hallucinations target turns dense in numerical data or financial jargon). Hallucinations are injected via minimal edits to key words or phrases, preserving the original response structure. We execute this pipeline using both GPT-4o and Gemini-2.5-Flash for the test split to mitigate model-specific artifacts, and GPT-4o only for the training split. The strategies are as follows:

\begin{itemize}
\item \textbf{Contradictory}: Induces subtle but critical deviations through four primary error types: (1) \textit{Financial Term Misunderstanding}, (2) \textit{Number Error}, (3) \textit{Modifier Change}, and (4) \textit{Inconsistency} with prior dialogue history.

\item \textbf{Unverifiable}: Inserts fabricated details or subjective opinions to ensure the response lacks evidentiary support from the context.

\item \textbf{Irrelevance}: Modifies the response to ignore query constraints or address a different aspect of the same topic.

\item \textbf{False Acceptance}: Replaces $\mathcal{P}_n$ with all hard negatives while retaining the original response, simulating model acceptance of an unanswerable query.

\item \textbf{False Refusal}: Replaces the response with a refusal string while keeping the original $\mathcal{P}_n$ intact.
\end{itemize}

Subtype-level injection strategies and prompt templates are provided in Appendix~\ref{appendix:prompt_templates}.

\paragraph{Quality Control}
For the training split, each injected turn is rated by an LLM judge on a 1--3 scale for \textit{appropriateness} (whether the intended hallucination type is correctly injected) and \textit{naturalness} (whether the injected response fits naturally in context), with turns below 2 on either dimension discarded. This filtering retains 85.2\% of train dialogues. We validate this judge against human annotations (Table~\ref{tab:faithful_dialogue_agreement}). For the test set, human annotators review and revise all injected turns without automated filtering. Detailed rubrics and annotation procedures are in Appendix~\ref{appendix:annotation_ui}.

\begin{table}[h]
  \centering
  \small
  \setlength{\tabcolsep}{4pt}
  \begin{tabular}{lcccc}
    \hline
    \multirow{2}{*}{\textbf{Metric}} & \multicolumn{2}{c}{\textbf{Human-Human}} & \multicolumn{2}{c}{\textbf{Human-LLM}} \\
    \cline{2-5}
    & \textbf{Agreement} & \textbf{AC1} & \textbf{Agreement} & \textbf{AC1} \\
    \hline
    \multicolumn{5}{l}{\textit{\textbf{Dialogue Quality}}} \\
    Answerability & 0.95 & 0.95 & 0.91 & 0.90 \\
    Plausibility  & 0.79 & 0.76 & 0.86 & 0.85 \\
    Correctness   & 0.96 & 0.96 & 0.97 & 0.96 \\
    Diversity     & 0.78 & 0.72 & 0.73 & 0.64 \\
    Coherency     & 0.75 & 0.70 & 0.70 & 0.64 \\
    \hline
    \multicolumn{5}{l}{\textit{\textbf{Injected Hallucination Quality}}} \\
    Appropriateness & 0.85 & 0.82 & 0.77 & 0.73 \\
    Naturalness     & 0.80 & 0.77 & 0.69 & 0.64 \\
    \hline
  \end{tabular}
  \caption{Agreement and Gwet's AC1 for the two quality-control stages: dialogue quality and injected hallucination quality. For each stage, four annotators each evaluate 30 of 40 sampled items (each spanning multiple turns), with three-way overlap per item.
  }
  \label{tab:faithful_dialogue_agreement}
\end{table}

\subsection{Dataset Statistics} \label{sec:dataset_stats}
\begin{figure*}[t]
  \includegraphics[width=0.98\linewidth]{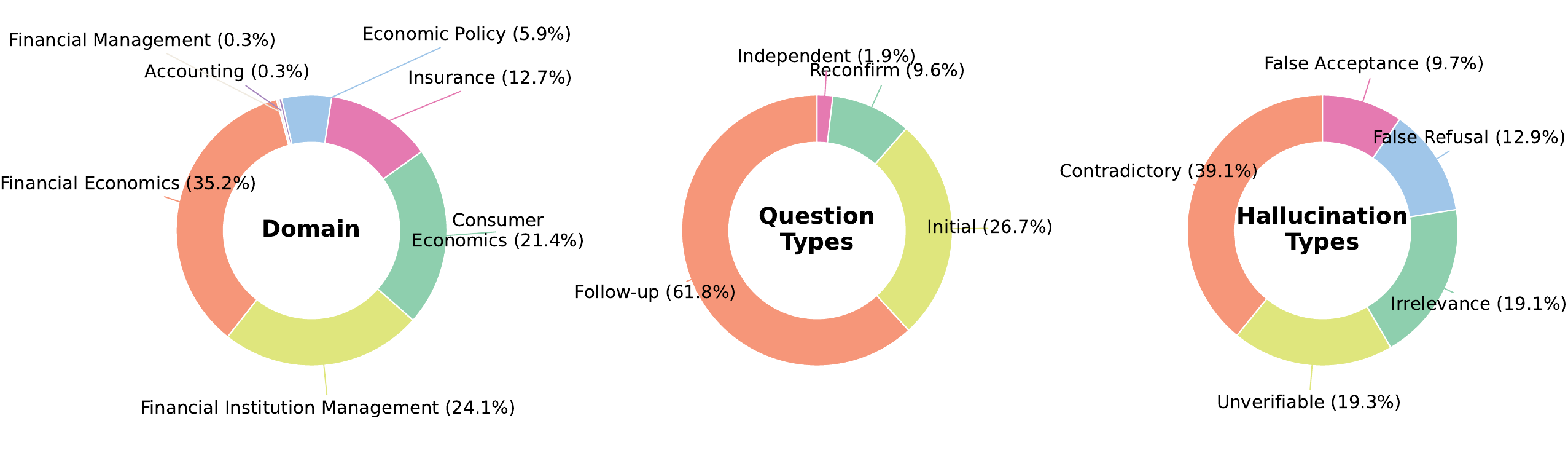}
  \caption {Distribution of K-FinHallu test set characteristics. The figure presents the percentage distributions of financial domains, question types, and hallucination types.}
  \label{fig:dataset_statistics}
  \vspace{-5pt}
\end{figure*}

K-FinHallu comprises a test split for evaluation and a training split for fine-tuning (Table~\ref{tab:dataset_stats}). The test split contains 808 evaluation samples, comprising 404 faithful and 404 hallucinated dialogues derived from 202 base dialogues. Each base dialogue contains between 4 and 8 turns, where exactly one hallucination is injected per base dialogue, distributed across turn positions to evaluate consistency throughout the conversation. About 9.7\% of samples present unanswerable scenarios to evaluate honest abstention. The training split contains 2{,}624 samples: 1{,}312 hallucinated dialogues paired with their faithful counterparts, via automated quality filtering. Figure~\ref{fig:dataset_statistics} presents the test-set distribution of financial domains, question types, and hallucination types, and Appendix~\ref{appendix:dataset_distribution} compares domain distributions across splits.

\begin{table}[t]
  \centering
  \small
  \begin{tabular}{lcc}
    \hline
    & \textbf{Test} & \textbf{Train} \\
    \hline
    \multicolumn{3}{l}{\textit{\textbf{Base Dialogues}}} \\
    Total \# of Dialogues & 202 & 1{,}312 \\
    \# of Turns (min--max) & 4--8 & 4--8 \\
    \hline
    \multicolumn{3}{l}{\textit{\textbf{K-FinHallu}}} \\
    Total \# of Dialogues & 808 & 2{,}624 \\
    Total \# of Unique Passages & 1{,}273 & 11{,}281 \\
    \# of Turns (min--max) & 1--8 & 1--8 \\
    Avg.\ \# of Turns & 3.72 & 3.61 \\
    Avg.\ \# of Question Words & 9.7 & 10.4 \\
    Avg.\ \# of Response Words & 16.1 & 18.1 \\
    Unanswerable (\%) & 9.7 & 11.4 \\
    \hline
  \end{tabular}
  \caption{Dataset statistics. Base Dialogues are faithful dialogues before hallucination injection. K-FinHallu combines hallucinated and faithful samples derived from these base dialogues.}
  \label{tab:dataset_stats}
\end{table}

\section{Experiments}

\subsection{Evaluation Setup and Metrics}
\label{sec:eval-setup}
We evaluate the detection task defined in Section~\ref{subsec:taskformulation} at the turn level to simulate real-time monitoring of responses within a RAG pipeline. The evaluation proceeds in two settings: \textbf{Binary classification} determines the presence of a hallucination, reporting accuracy, precision, recall, and F1. \textbf{Four-class classification} evaluates answerability judgment and refusal behavior across \textit{Faithful Answer}, \textit{Hallucination} (covering \textit{Contradictory}, \textit{Unverifiable}, \textit{Irrelevance}, and \textit{False Acceptance}), \textit{False Refusal}, and \textit{True Refusal}, reporting per-class accuracy. For readability, we collapse \textit{False Refusal} and \textit{True Refusal} into a single \textit{Refusal} class.

\subsection{Baselines}

To detect hallucination, we select various LLMs that have shown strong performance in the Korean NLP community and are widely used in research or industry: (1) \textit{Korean-centric} models (kanana-2-30b-a3b-instruct~\cite{kanana}, EXAONE-4.0-32B~\cite{exaone-4.0}), (2) \textit{open-source} models (Llama-3.1 (8B, 70B), Llama-3.3 (70B)~\cite{llama3.1}, Qwen3 series~\cite{qwen3technicalreport}), and (3) \textit{closed-source} models (Gemini-2.5 (Flash, Pro)~\cite{gemini25pushingfrontier}, GPT-4o~\cite{openai2024gpt4ocard}, GPT-5~\cite{openai2025gpt5systemcard}). For reproducibility, all models are evaluated at a temperature of 0 where applicable.

In addition to these off-the-shelf baselines, we fine-tune Qwen3-8B on the training split using LoRA~\cite{hu2022lora} for the four-class detection task. We train two variants: \textbf{Qwen3-8B SFT} is trained to predict only the class label, whereas \textbf{Qwen3-8B SFT-R} is additionally supervised with a templated rationale that mirrors our taxonomy's two-step reasoning: assessing answerability, then evaluating the response (Appendix~\ref{appendix:sft}).

\subsection{Main Results}

\begin{table}[t]
  \centering
  \small
  \begin{tabular}{lcccc}
    \hline
    \textbf{Model} & \textbf{Acc.} & \textbf{Prec.} & \textbf{Rec.} & \textbf{F1} \\
    \hline
    \multicolumn{5}{l}{\textit{\textbf{Korean-centric}}} \\
    kanana-2-30b-a3b & 0.533 & 0.520 & 0.866 & 0.650 \\
    EXAONE-4.0-32B & 0.717 & 0.700 & 0.757 & 0.728 \\
    \hline
    \multicolumn{5}{l}{\textit{\textbf{Open-source}}} \\
    Llama-3.1-8B & 0.537 & 0.520 & \textbf{0.958} & 0.674 \\
    Llama-3.1-70B & 0.700 & 0.659 & 0.832 & 0.735 \\
    Llama-3.3-70B & 0.722 & 0.672 & 0.864 & 0.756 \\
    Qwen3-8B & 0.720 & 0.850 & 0.533 & 0.655 \\
    Qwen3-14B & 0.756 & 0.881 & 0.590 & 0.707 \\
    Qwen3-32B & 0.775 & \underline{0.905} & 0.614 & 0.732 \\
    \hline
    \multicolumn{5}{l}{\textit{\textbf{Closed-source}}} \\
    GPT-4o & 0.715 & 0.691 & 0.780 & 0.733 \\
    GPT-5 & 0.833 & 0.877 & 0.775 & 0.823 \\
    Gemini-2.5-Flash & \textbf{0.864} & 0.883 & \textbf{0.839} & \textbf{0.860} \\
    Gemini-2.5-Pro & \underline{0.859} & \textbf{0.912} & 0.795 & \underline{0.849} \\
    \hline
  \end{tabular}
  \caption{Performance of evaluated LLMs on binary hallucination detection. Metrics include Accuracy (Acc.), Precision (Prec.), Recall (Rec.), and F1-score (F1). Bold and underlined values denote the best and second-best scores per column.}
  \label{tab:binary-eval}
\end{table}

\begin{table}[t]
  \centering
  \small
  \begin{tabular}{lcccc}
    \hline
    \textbf{Model} & \textbf{Faith.} & \textbf{Hallu.} & \textbf{Ref.} & \textbf{Overall} \\
    \hline
    \multicolumn{5}{l}{\textit{\textbf{Korean-centric}}} \\
    kanana-2-30b-a3b & 0.597 & 0.494 & 0.308 & 0.515 \\
    EXAONE-4.0-32B & 0.727 & 0.514 & 0.471 & 0.601 \\
    \hline
    \multicolumn{5}{l}{\textit{\textbf{Open-source}}} \\
    Llama-3.1-8B & 0.114 & 0.537 & 0.452 & 0.342 \\
    Llama-3.1-70B & 0.781 & 0.670 & 0.471 & 0.693 \\
    Llama-3.3-70B & 0.756 & 0.682 & 0.471 & 0.687 \\
    Qwen3-8B & \textbf{0.986} & 0.446 & 0.490 & 0.688 \\
    Qwen3-14B & 0.977 & 0.554 & 0.567 & 0.741 \\
    Qwen3-32B & 0.969 & 0.551 & 0.625 & 0.743 \\
    \hline
    \multicolumn{5}{l}{\textit{\textbf{Closed-source}}} \\
    GPT-4o & 0.835 & 0.642 & 0.375 & 0.692 \\
    GPT-5 & 0.969 & 0.810 & 0.625 & \underline{0.855} \\
    Gemini-2.5-Flash & 0.946 & 0.795 & 0.673 & 0.845 \\
    Gemini-2.5-Pro & 0.940 & \underline{0.815} & \underline{0.683} & 0.853 \\
    \hline
    \multicolumn{5}{l}{\textit{\textbf{Fine-tuned (ours)}}} \\
    Qwen3-8B SFT & \underline{0.983} & 0.764 & 0.433 & 0.822 \\
    Qwen3-8B SFT-R & 0.972 & \textbf{0.864} & \textbf{0.750} & \textbf{0.896} \\
    \hline
  \end{tabular}
  \caption{\label{tab:fourclass-acc} Per-class accuracy on the four-class detection task under the three-class collapsed view. \textbf{Faith.}: Faithful. \textbf{Hallu.}: Hallucinated. \textbf{Ref.}: Refusal (False + True). \textit{Overall} is the micro-average. Full breakdown in Table~\ref{tab:fourclass-full}.}
\end{table}

\paragraph{Detection Performance}
Binary detection results are summarized in Table~\ref{tab:binary-eval}. Closed-source models lead overall, with Gemini-2.5-Flash achieving the highest F1 of 0.860, followed by Gemini-2.5-Pro (0.849) and GPT-5 (0.823). Among open-source models, Qwen3-32B (0.732) is comparable to GPT-4o (0.733), with consistent scaling gains within the Qwen3 family. Korean-centric models lag behind their global counterparts of similar size.
A clear precision--recall trade-off distinguishes model families: Llama-3.1-8B exhibits a ``pessimistic bias,'' with the highest recall (0.958) but the lowest precision (0.520), suggesting a tendency to over-identify faithful responses as hallucinations. The Qwen3 series shows the opposite pattern, favoring precision at the expense of recall.

\paragraph{Closing the Gap with Fine-tuning}
Table~\ref{tab:fourclass-acc} reports the four-class results under a three-class collapsed view. The three axes exhibit distinct difficulty levels: Faithful classification is near-ceiling for most models, Hallucination detection shows the largest inter-model variance, and Refusal scores below 0.50 for most base models.
Overall, GPT-5 attains the highest accuracy among base models (0.855), with the Gemini-2.5 series close behind. Open-source base models trail substantially, with Qwen3-32B the strongest at 0.743.

Fine-tuning Qwen3-8B with rationale supervision (Qwen3-8B SFT-R) achieves an overall accuracy of 0.896, outperforming all base models. The largest gain over the base Qwen3-8B is on Hallucination detection, where accuracy improves from 0.446 to 0.864, exceeding GPT-5. Notably, rationale supervision is critical, as Qwen3-8B SFT without rationale scores 0.822.

Although Qwen3-8B SFT-R achieves the highest Refusal accuracy at 0.750, Refusal remains the weakest axis across all models. Separating \textit{True Refusal} from \textit{False Refusal} (Table~\ref{tab:fourclass-full}) reveals two failure patterns among base models: open-source and Korean-centric models fail to distinguish between the two refusal types, predicting one almost exclusively. Closed-source models detect \textit{False Refusal} accurately but struggle to recognize \textit{True Refusal} as a valid abstention, treating refusal itself as a negative signal.

\subsection{Analysis}

\paragraph{Multi-class Diagnostics}
To further probe diagnostic capability, we evaluate multi-class classification on the hallucinated samples to identify the underlying cause among \textit{Contradictory}, \textit{Unverifiable}, \textit{Irrelevance}, \textit{False Acceptance}, and \textit{False Refusal}. Results (Table~\ref{tab:multiclass-f1}) reveal a substantial drop from binary detection, confirming that diagnosing the \textit{cause} of a hallucination is harder than detecting its \textit{presence}. The Qwen3 series follows a clear scaling trend where performance improves with model size (0.387 to 0.492 macro-F1 from 8B to 32B), while Korean-centric models and GPT-4o tend to collapse into a single dominant type, scoring high only in specific categories while failing in others. Notably, confusion between \textit{Contradictory} and \textit{Unverifiable} persists even for GPT-5 (confusion matrices in Table~\ref{tab:cm-gpt5}).

\begin{figure}[t]
  \includegraphics[width=0.9\columnwidth]{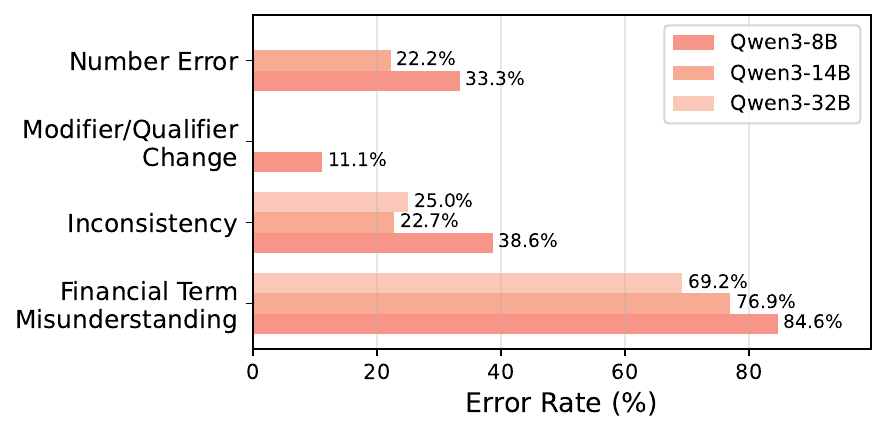}
  \caption{Error rates of the Qwen3 series on \textit{Contradictory} sub-types (GPT-4o-injected subset). Each bar represents the proportion of hallucinations undetected by the model.}
  \label{fig:qwen_contradictory}
  \vspace{-5pt}
\end{figure}

\paragraph{Contradictory Sub-type Analysis}
We further decompose \textit{Contradictory} hallucinations, the most frequent type in our benchmark, into four sub-causes. This analysis uses the GPT-4o-injected subset where sub-type labels are manually annotated.
As shown in Figure~\ref{fig:qwen_contradictory}, error rates for \textit{Number Error} and \textit{Modifier Change} decrease with model scale in the Qwen3 series, but \textit{Financial Term Misunderstanding} shows minimal improvement even at 32B. This suggests that effective detection in the financial domain requires nuanced domain expertise beyond surface-level lexical matching.

\paragraph{Cross-source Analysis}
K-FinHallu employs both GPT-4o and Gemini-2.5-Flash as injection sources over the same faithful dialogues for the test split to mitigate model-specific artifacts (Table~\ref{tab:cross-source}). The relative ordering of strong and weak models is largely preserved across both subsets, but the two sources differ in difficulty: for most hallucination types the gap is modest ($\pm$10\%p), but \textit{Irrelevance} shows a pronounced difference. GPT-4o produces topic-level drift that is readily identifiable, whereas Gemini-2.5-Flash substitutes adjacent information from the same document that appears relevant but does not answer the query (e.g., Gemini-2.5-Pro drops from 0.93 to 0.68). This variation in difficulty demonstrates that source diversity strengthens evaluation robustness by exposing different failure modes.

\section{Conclusion}
We introduced K-FinHallu, a multi-turn RAG hallucination detection benchmark specialized for the Korean financial domain. By establishing a hierarchical taxonomy based on context answerability and implementing a hallucination injection pipeline using authentic financial documents, we benchmarked frontier and open-source LLMs as hallucination detectors. Our results demonstrate that even frontier models struggle with answerability judgment and fine-grained diagnosis, despite competent binary detection, highlighting justified abstention as a critical yet under-evaluated dimension in financial RAG. In particular, \textit{Financial Term Misunderstanding} persists as a primary bottleneck regardless of model scale, suggesting that even frontier LLMs struggle to capture the semantic shifts caused by subtle alterations in Korean financial jargon. A fine-tuned 8B model with rationale supervision achieves competitive or superior performance to frontier baselines, offering a practical path to open-source detector development. We release K-FinHallu under the CC-BY-NC 4.0 license to support further research in this direction.

\section*{Limitations}

Our study is subject to several limitations. First, while K-FinHallu is curated from authoritative Korean financial sources, its coverage remains focused on core domains and may underrepresent highly specialized sub-sectors such as complex derivatives or investment banking. Second, we inject exactly one hallucination per dialogue by design to support precise, turn-level diagnosis under our real-time monitoring setting. This controlled setup does not capture co-occurring or compounding hallucinations that may arise in real-world interactions, which we leave to future work. Third, the dialogues in K-FinHallu were generated via LLM simulation, a process that may pass down generator-specific linguistic patterns or artifacts to the dataset, though all five hallucination types were also observed in naturally generated responses from frontier LLMs. Finally, our terminology-related findings are grounded in Korean regulatory language and market context, so transferability to other languages or jurisdictions may be limited.

\section*{Acknowledgments}
This paper used datasets from `The Open AI Dataset Project (AI-Hub, S.~Korea)'. All data information can be accessed through `AI-Hub (www.aihub.or.kr)'.


\bibliography{custom}

\appendix

\section{Real-world Validation of Hallucination Taxonomy}
\label{appendix:realism}

K-FinHallu uses controlled hallucination injection to enable systematic and balanced evaluation. A natural concern, however, is whether the resulting benchmark (both its taxonomy and the injected examples) reflects hallucinations encountered in real deployment. We address this through three lines of empirical evidence: (i) the natural occurrence rate and distribution of hallucinations from frontier LLMs under free generation, (ii) aggregate statistics from a real-world RAG deployment, and (iii) qualitative differences across source models used for injection.

\subsection{Natural Hallucination Distribution from Frontier LLMs}
\label{appendix:realism_natural}

To assess whether our taxonomy reflects actual model hallucination patterns, GPT-4o and Gemini-2.5-Flash generate a response for each of the 404 faithful samples in K-FinHallu (temperature 0.7), yielding 808 model responses in total. Each response is then labeled by a separate LLM judge (GPT-5-mini) against the original answer and its clue text, with refusal cases detected by keyword matching.

Of the 808 responses, 282 are classified as hallucinations. The majority (247) belong to refusal-related types, comprising 189 \textit{False Refusal} and 58 \textit{False Acceptance} cases. Content-based hallucinations (\textit{Contradictory}, \textit{Unverifiable}, \textit{Irrelevance}) account for only 35 cases, with GPT-4o producing 14 (3.5\%) and Gemini-2.5-Flash producing 21 (5.2\%). The per-model totals are 214 for GPT-4o and 68 for Gemini-2.5-Flash.

Despite this low rate, all five hallucination types in our taxonomy are observed in these frontier LLMs (Table~\ref{tab:natural-distribution}). The two models, however, exhibit distinct biases: GPT-4o concentrates on \textit{False Refusal} (186 of 214), while Gemini-2.5-Flash concentrates on \textit{False Acceptance} (44 of 68). Since each model has different weak spots and natural generation alone produces a highly skewed distribution across the five types, controlled injection is required for balanced evaluation, especially given the high-stakes nature of financial RAG.

\begin{table}[h]
\centering
\small
\begin{tabular}{lccc}
\toprule
\textbf{Type} & \textbf{GPT-4o} & \textbf{Gemini-2.5-Flash} & \textbf{Total} \\
\midrule
Contradictory     & 9   & 10 & 19  \\
Unverifiable      & 4   & 9  & 13  \\
Irrelevance       & 1   & 2  & 3   \\
False Refusal     & 186 & 3  & 189 \\
False Acceptance  & 14  & 44 & 58  \\
\midrule
Total             & 214 & 68 & 282 \\
\bottomrule
\end{tabular}
\caption{Distribution of hallucinations naturally produced by GPT-4o and Gemini-2.5-Flash on the 404 K-FinHallu faithful samples (temperature 0.7, judged by GPT-5-mini).}
\label{tab:natural-distribution}
\end{table}

\subsection{Real-world RAG Deployment Statistics}
\label{appendix:realism_fgt}

We additionally analyze hallucination prevalence in a real RAG-based financial service through an internal field test of a deployed system. Among 2{,}265 RAG-generated responses evaluated by human assessors, 960 were flagged as errors (a broad category including hallucination, inappropriate handling of prompt attacks, answers requiring trimming, and outdated reference documents). Of these flagged errors, 434 (approximately 45\%) underwent detailed manual inspection, and 224 (51.6\%) were confirmed as hallucinations. Extrapolating to the full set yields an estimated hallucination rate of approximately 21.9\% in deployment.

The per-type breakdown (see Table~\ref{tab:fgt-distribution}) indicates that the types defined in our taxonomy are observed in real deployment, suggesting that the taxonomy reflects practically relevant failure modes. The per-type ratios differ from our balanced injection scheme, and the distribution in deployment may shift with service settings; we therefore view our balanced injection as a controlled stress test rather than a faithful reproduction of any single operational distribution.

\begin{table*}[h]
\centering
\small
\begin{tabular}{llc}
\toprule
\textbf{Type (\%)} & \textbf{Subtype} & \textbf{Count} \\
\midrule
\multirow{4}{*}{\shortstack[l]{Contradictory\\(20.09\%)}}
 & Inconsistency                   & 11 \\
 & Financial Term Misunderstanding & 11 \\
 & Modifier Change                 & 1  \\
 & Number Error                    & 22 \\
\midrule
\multirow{2}{*}{\shortstack[l]{Unverifiable\\(36.61\%)}}
 & Fabricated Detail                & 73 \\
 & Subjective Opinion              & 9  \\
\midrule
\multirow{3}{*}{\shortstack[l]{Irrelevance\\(9.37\%)}}
 & Condition Not Satisfied          & 10 \\
 & Question Misunderstanding        & 3  \\
 & Off-topic Drift                  & 8  \\
\midrule
False Refusal (4.91\%)     & ---  & 11 \\
False Acceptance (29.02\%) & ---  & 65 \\
\midrule
\textbf{Total (100.00\%)}  &      & \textbf{224} \\
\bottomrule
\end{tabular}
\caption{Per-type and per-subtype distribution of hallucinations identified from a pilot study of a RAG-based financial chatbot ($n=224$ hallucinations out of 434 manually inspected error cases). Type-level percentages (shown next to each type name) are computed over the 224 confirmed hallucinations.}
\label{tab:fgt-distribution}
\end{table*}

The data was collected and analyzed in collaboration with financial domain experts at an industry partner operating a deployed RAG-based financial service. Hallucination judgments were made by domain experts using a shared rubric, and category-level labels were further verified through manual inspection. The raw cases involve proprietary user interactions and cannot be released; only aggregate statistics are reported here.

\subsection{Cross-source Characteristics of Injected Hallucinations}
\label{appendix:realism_cross_source}

For the test split, we use both GPT-4o and Gemini-2.5-Flash as injection sources to mitigate source-model-specific artifacts. Detector accuracy on the two subsets is reported quantitatively in Table~\ref{tab:cross-source}, and we additionally observe systematic qualitative differences between the two sources, particularly for \textit{Irrelevance}. GPT-4o-injected \textit{Irrelevance} tends to produce clearly off-topic responses, whereas Gemini-injected \textit{Irrelevance} substitutes adjacent information from the same context, making detection considerably harder. For instance, when asked when the monitoring of MMF mark-to-market preparation begins, a Gemini-injected response cites the regulation's effective date drawn from the same timeline. This information lies within the source context but answers a different question. These cross-source differences suggest that the characteristics of injected hallucinations depend on the source model, and that source-model diversity may be a useful axis for improving benchmark robustness.

\section{Dataset Distribution Across Splits}
\label{appendix:dataset_distribution}

The corpus is partitioned by source institution to prevent data leakage (Section~\ref{sec:dataset_stats}). The test split is drawn from the Financial Supervisory Service ($\sim$72\%) and the Korea Consumer Agency ($\sim$28\%). The training split is drawn primarily from the Bank of Korea and the Financial Services Commission, with smaller contributions from the Korea Institute of Finance and the Ministry of Employment and Labor.

\begin{figure*}[t]
  \centering
  \includegraphics[width=0.95\linewidth]{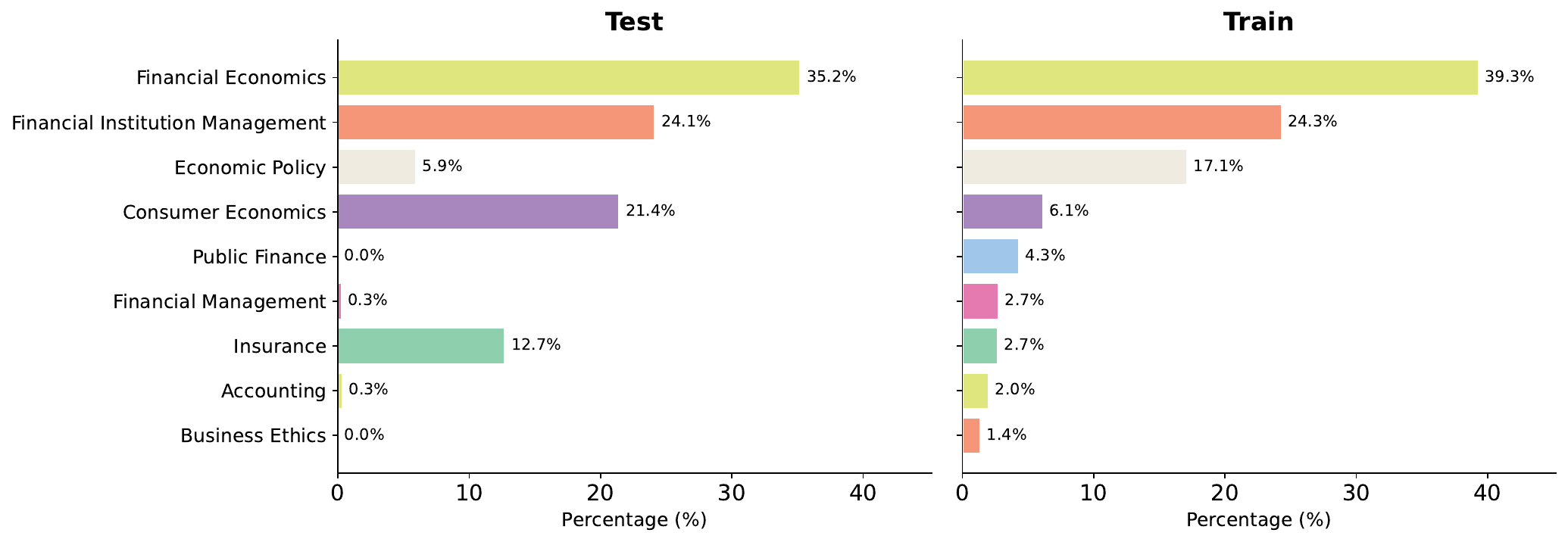}
  \caption{Document domain distribution in the test (left) and training (right) splits. The same domain is assigned a consistent color across both panels.}
  \label{fig:domain_test_vs_train}
\end{figure*}

As shown in Figure~\ref{fig:domain_test_vs_train}, both splits are dominated by \textit{Financial Economics} and \textit{Financial Institution Management}. The training split additionally covers \textit{Economic Policy} (17.1\%), \textit{Public Finance}, and \textit{Business Ethics}, which are sparse or absent in the test split, while the test split is more concentrated in \textit{Consumer Economics} (21.4\%) and \textit{Insurance} (12.7\%). These differences reflect the distinct subject coverage of the source institutions.

\section{Retrieval Simulation}
\label{appendix:passage_classifier}

To rigorously evaluate context adherence, K-FinHallu constructs $\mathcal{P}_n$ to include both \textit{answerable} and \textit{unanswerable} contexts. For each query, the top-$k$ candidate passages are retrieved using the KURE embedding model~\cite{KURE} based on cosine similarity with the decontextualized query $\tilde{q}_n$, which mitigates coreference and ellipsis in the conversational utterance $q_n$, following prior multi-turn RAG work~\cite{zhou2023unified, sun2023improving, katsis2025mtrag}. Following the notation in Section~\ref{sec:dataset_stats}, each retrieved passage falls into one of three types:
\begin{itemize}\itemsep0pt
    \item \textbf{gold passage} $p^*$: the direct source of the response, containing the grounding clue $c_n$.
    \item \textbf{positive passage} $p^+$: not the gold passage, but still supports the answer through paraphrase or topical overlap with $p^*$.
    \item \textbf{hard negative} $p^-$: semantically similar to $\tilde{q}_n$ but contains neither the answer nor its grounding clue.
\end{itemize}
\textit{Answerable} contexts always include $p^*$ within the top-$k$, while \textit{unanswerable} contexts consist solely of $p^-$. However, similarity-based retrieval alone cannot prevent $p^+$ from mixing into the unanswerable candidate pool.

To filter out positive passages, we apply a two-stage LLM-based classifier to the retrieved candidates in descending order of similarity. The first stage, \textit{context sufficiency}, evaluates the input pair $(\tilde{q}_n, p_n)$ to determine whether the query $q_n$ is \textit{answerable} based solely on $p_n$, without relying on external knowledge. The second stage, \textit{clue presence}, assesses the pair $(c_n, p_n)$ by checking whether the clue text $c_n$ is explicitly mentioned or logically supported within $p_n$. For each stage, the classifier produces a binary judgment (yes/no) alongside a natural language reasoning trace.

Only candidates receiving a ``no'' judgment in both stages are accepted as hard negatives $p^-$, and the procedure continues sequentially until enough hard negatives are collected. The selected hard negatives are then exhaustively inspected by human annotators to confirm that no positive passages remain. This controlled manipulation enables systematic evaluation of honest abstention and hallucination robustness under both answerable and unanswerable retrieval contexts.

\section{Human Annotation}
\label{appendix:annotation_ui}
A total of four annotators participated in the annotation process: two graduate students in NLP and two financial domain experts. For each item, three annotators were randomly assigned and conducted data verification under a shared set of guidelines. The annotators assessed both faithful dialogues and hallucination-injected samples in a systematic manner. Following the initial evaluation, additional manual refinement was conducted as a quality control to ensure that each query and response accurately reflected the intended hallucination type and that the overall dialogue remained natural and coherent.

\subsection{Faithful Dialogue Evaluation}
The faithfulness of constructed dialogues is assessed along five dimensions proposed by \cite{lee2024multi}: \textit{answerability}, \textit{plausibility}, \textit{correctness}, \textit{diversity}, and \textit{coherency}. These criteria are applied at the query $q_n$, response $a_n$, and dialogue levels to capture both turn-level validity and overall dialogue consistency.

\begin{figure*}[t]
  \includegraphics[width=0.98\linewidth]{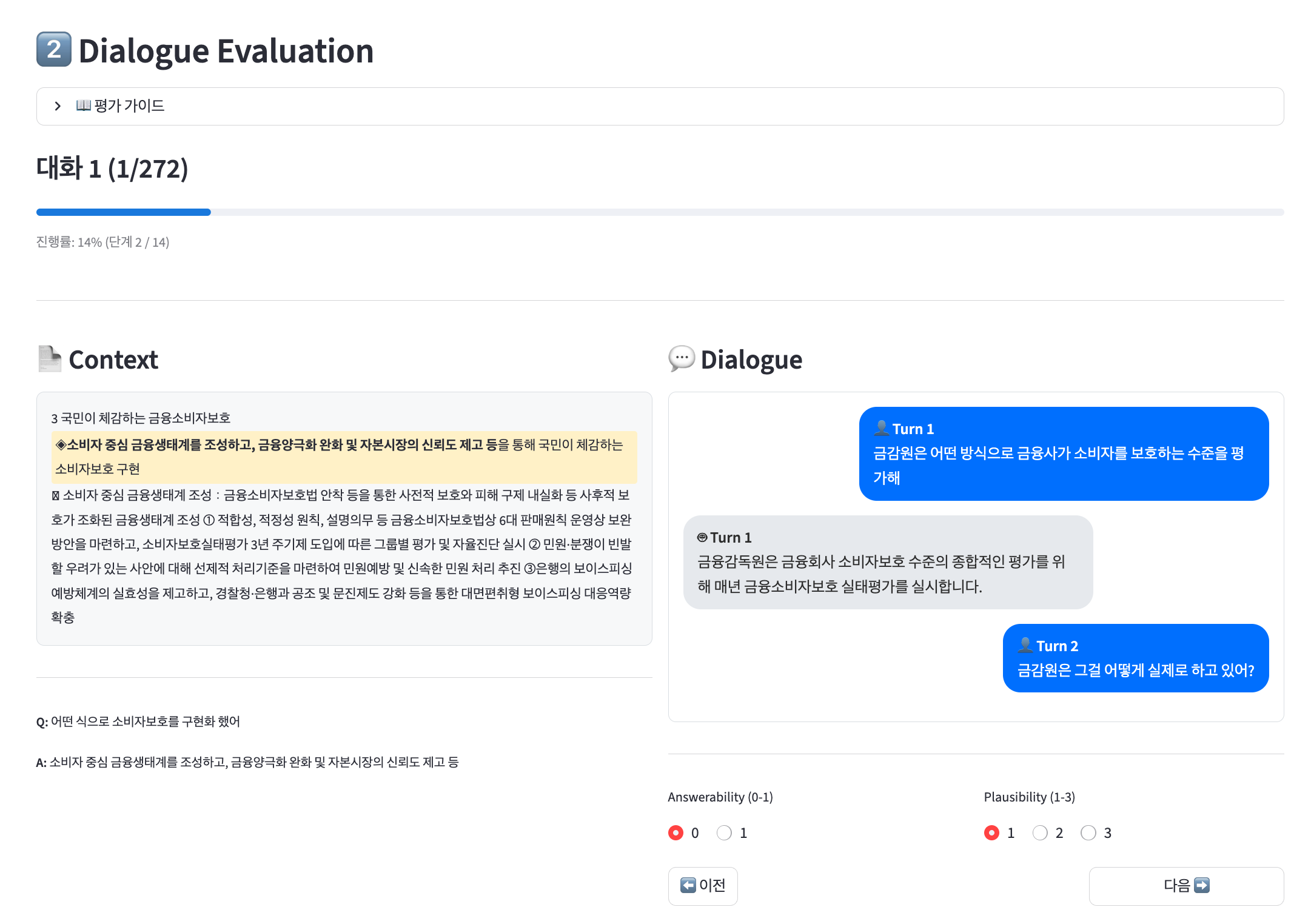}
  \caption {Screenshot of the annotation interface for dataset quality control.}
  \label{fig:annot}
\end{figure*}

At the query level, \textit{answerability} evaluates whether a query can be reasonably answered given the provided context, while \textit{plausibility} assesses whether the query is consistent with realistic user behavior. At the response level, \textit{correctness} measures factual accuracy and passage grounding. At the dialogue level, \textit{diversity} ensures that successive turns introduce non-redundant information, and \textit{coherency} verifies that each turn appropriately reflects prior context and maintains logical consistency. Overall, these metrics provide a comprehensive characterization of faithful multi-turn dialogues.
To facilitate this multi-dimensional assessment, we developed a dedicated annotation interface that presents the dialogue history, retrieved context, and candidate responses, as depicted in Figure~\ref{fig:annot}.

\begin{table*}[t]
\centering
\small
\setlength{\tabcolsep}{8pt}
\begin{tabular}{@{}p{1.5cm}p{3cm}p{10.5cm}@{}}
\toprule
\textbf{Target} & \textbf{Criteria} & \textbf{Description and Scoring Rubric} \\ \midrule

\multirow{2}{*}{\textbf{Query}} & \textbf{Answerability} (0--1) &
\textbf{Description}: Is the answer to the query present in the given documents? \newline
\textbf{0}: The document does not contain the information needed, or it cannot be derived. \newline
\textbf{1}: The answer is explicitly stated or can be reasonably inferred from the document. \\ \cmidrule{2-3}

 & \textbf{Plausibility} (1--3) &
\textbf{Description}: How likely is the query created by an actual user? \newline
\textbf{1}: Clearly unnatural, contrived, or unlikely to be asked by a real user. \newline
\textbf{2}: Generally plausible, but slightly awkward or artificial in phrasing. \newline
\textbf{3}: Natural and realistic; highly likely to be asked by a real user. \\ \midrule

\textbf{Response} & \textbf{Correctness} (0--1) & 
\textbf{Description}: Is the response correct based on the provided documents? \newline
\textbf{0}: Contains incorrect claims, contradictions, or unsupported information. \newline
\textbf{1}: Fully consistent with the document and answers the question correctly. \\ \midrule

\multirow{2}{*}{\textbf{Dialogue}} & \textbf{Diversity} (1--3) & 
\textbf{Description}: Is each query-response in the dialog sufficiently different? \newline
\textit{*Exclusion}: One intentionally included verification turn is excluded from scoring. \newline
\textbf{1}: Largely repetitive in wording, structure, or content. \newline
\textbf{2}: Adds some new info, but sentence structure remains noticeably monotonous. \newline
\textbf{3}: Each turn adds clearly new information with varied wording and structure. \\ \cmidrule{2-3}

 & \textbf{Coherency} (1--3) &
\textbf{Description}: Is the dialog flow natural and coherent? \newline
\textbf{1}: Incoherent; irrelevant to prior context or contains logical contradictions. \newline
\textbf{2}: Partially coherent; related to context but has awkward transitions. \newline
\textbf{3}: Coherent and consistent; accurately reflects context with a natural flow. \\ \bottomrule
\end{tabular}
\caption{Evaluation rubric for dialogue quality, adapted from \cite{lee2024multi}. Metrics are categorized by their evaluation targets: query, response, and overall dialogue.}
\label{tab:dialogue_eval_rubric}
\end{table*}

\subsection{Hallucinated Dialogue Evaluation}
Hallucinated dialogues are evaluated on two criteria, each rated on a 1--3 scale: \textit{appropriateness}, which measures whether the intended hallucination type is correctly exhibited, and \textit{naturalness}, which assesses whether the response maintains fluency and coherence within the dialogue context. Each response carries a single hallucination type assigned during injection. Annotators verify this label and flag mismatches for revision or removal.

\section{Fine-tuning Setup}
\label{appendix:sft}

We fine-tune Qwen3-8B on the training split of K-FinHallu (2{,}624 samples) for the four-class detection task introduced in Section~\ref{sec:taxonomy}. The training split is partitioned 9:1 into training and validation subsets.

\paragraph{Optimization.}
We fine-tune using the ms-swift framework~\cite{swift} with LoRA~\cite{hu2022lora} at rank $r{=}8$, $\alpha{=}32$, and a dropout of 0.05, attached to all linear projections of the backbone. We train for 3 epochs with a learning rate of $1\mathrm{e}{-4}$, a cosine schedule, and a 3\% warmup ratio. The effective batch size is 32 (per-device batch 2, gradient accumulation 2, across 8 GPUs). Training runs on 8 $\times$ NVIDIA RTX A6000 (48~GB) with bfloat16 mixed precision, completing in approximately 4 hours.

\paragraph{Inference.}
All open-source baselines and the fine-tuned models are served with vLLM~\cite{kwon2023efficient} using tensor parallelism across 4 NVIDIA RTX A6000 (48~GB) GPUs. Closed-source models are accessed via their official APIs. All evaluations use a temperature of 0 where applicable for reproducibility.

\paragraph{Rationale template.}
For the \textit{Qwen3-8B SFT-R} variant, each training instance is paired with a templated rationale that follows our taxonomy's two-step reasoning: first assessing answerability based on the presence of evidentiary clue text, then evaluating whether the response is faithful (when answerable) or correctly abstains (when unanswerable). The rationale is placed inside Qwen3's native \texttt{<think>} block so that the model is supervised to produce it during its thinking phase before emitting the class label. Training uses Korean templates. We provide their English translations below for readability.
\begin{itemize}\itemsep0pt
    \item \textbf{Faithful Answer}: ``The reference document contains the clue \{\textit{clue}\}, so the query is answerable. The current response answers it appropriately, therefore \textit{Faithful Answer}.''
    \item \textbf{Hallucination} (answerable case): ``The reference document contains the clue \{\textit{clue}\}, so the query is answerable. However, the current response provides an inappropriate answer, therefore \textit{Hallucination}.''
    \item \textbf{Hallucination} (unanswerable case): ``The reference document does not contain evidence to support the query, so the query is unanswerable. However, the current response provides an answer, therefore \textit{Hallucination}.''
    \item \textbf{False Refusal}: ``The reference document contains the clue \{\textit{clue}\}, so the query is answerable. However, the current response refuses to answer despite sufficient evidence, therefore \textit{False Refusal}.''
    \item \textbf{True Refusal}: ``The reference document does not contain evidence to support the query, so the query is unanswerable. The current response correctly refuses to answer, therefore \textit{True Refusal}.''
\end{itemize}
The \{\textit{clue}\} slot is filled with the evidentiary span $\tilde{c}$ from the source annotation when available. The \textit{Qwen3-8B SFT} (no reasoning) variant is trained on the same data with the rationale field removed.

\section{Additional Experimental Results}
\label{appendix:additional_results}

This appendix provides additional breakdowns and analyses that supplement the main results: the full four-class breakdown, multi-class diagnostics with confusion analysis, and cross-source robustness.

\subsection{Four-class Full Breakdown}
\label{appendix:fourclass-full}

Table~\ref{tab:fourclass-full} reports the full four-class breakdown without collapsing \textit{False Refusal} and \textit{True Refusal}. The breakdown surfaces two failure patterns among base models: collapse onto a single refusal type (0.000 on \textit{False Refusal}), and low \textit{True Refusal} among closed-source models.

\begin{table*}[t]
  \centering
  \small
  \begin{tabular}{lccccc}
    \hline
    \textbf{Model} & \textbf{Faithful} & \textbf{Hallucination} & \textbf{False Refusal} & \textbf{True Refusal} & \textbf{Overall} \\
    \hline
    \multicolumn{6}{l}{\textit{\textbf{Korean-centric}}} \\
    kanana-2-30b-a3b & 0.597 & 0.494 & 0.000 & 0.615 & 0.515 \\
    EXAONE-4.0-32B & 0.727 & 0.514 & 0.000 & \textbf{0.942} & 0.601 \\
    \hline
    \multicolumn{6}{l}{\textit{\textbf{Open-source}}} \\
    Llama-3.1-8B & 0.114 & 0.537 & 0.577 & 0.327 & 0.342 \\
    Llama-3.1-70B & 0.781 & 0.670 & 0.000 & \textbf{0.942} & 0.693 \\
    Llama-3.3-70B & 0.756 & 0.682 & 0.000 & \textbf{0.942} & 0.687 \\
    Qwen3-8B & \textbf{0.986} & 0.446 & 0.500 & 0.481 & 0.688 \\
    Qwen3-14B & 0.977 & 0.554 & 0.423 & 0.712 & 0.741 \\
    Qwen3-32B & 0.969 & 0.551 & 0.654 & 0.596 & 0.743 \\
    \hline
    \multicolumn{6}{l}{\textit{\textbf{Closed-source}}} \\
    GPT-4o & 0.835 & 0.642 & 0.096 & 0.654 & 0.692 \\
    GPT-5 & 0.969 & 0.810 & \textbf{0.904} & 0.346 & \underline{0.855} \\
    Gemini-2.5-Flash & 0.946 & 0.795 & \underline{0.808} & 0.538 & 0.845 \\
    Gemini-2.5-Pro & 0.940 & \underline{0.815} & 0.769 & 0.596 & 0.853 \\
    \hline
    \multicolumn{6}{l}{\textit{\textbf{Fine-tuned (ours)}}} \\
    Qwen3-8B SFT & \underline{0.983} & 0.764 & 0.173 & 0.692 & 0.822 \\
    Qwen3-8B SFT-R & 0.972 & \textbf{0.864} & 0.654 & \underline{0.846} & \textbf{0.896} \\
    \hline
  \end{tabular}
  \caption{\label{tab:fourclass-full} Per-class accuracy on the four-class detection task (full view, without collapsing False Refusal and True Refusal). \textit{Overall} is the micro-average across all 808 test samples. Bold and underlined values denote the best and second-best scores per column.}
\end{table*}

\subsection{Multi-class Classification Results}
\label{appendix:multiclass}

Table~\ref{tab:multiclass-f1} reports multi-class classification F1 over the 404 hallucinated samples. Korean-centric and Llama series models score near zero, the Qwen3 series shows moderate performance, while GPT-5 and the Gemini-2.5 series cluster around 0.70.

\begin{table*}[t]
  \centering
  \small
  \begin{tabular}{lcccccc}
    \hline
    \textbf{Model} & \textbf{Contradictory} & \textbf{Unverifiable} & \textbf{Irrelevance} & \textbf{False Acc.} & \textbf{False Ref.} & \textbf{Overall} \\
    \hline
    \multicolumn{7}{l}{\textit{\textbf{Korean-centric}}} \\
    kanana-2-30b-a3b & 0.000 & 0.000 & 0.000 & 0.109 & 0.000 & 0.022 \\
    EXAONE-4.0-32B & 0.000 & 0.049 & 0.000 & 0.146 & 0.000 & 0.039 \\
    \hline
    \multicolumn{7}{l}{\textit{\textbf{Open-source}}} \\
    Llama-3.1-8B & 0.024 & 0.205 & 0.000 & 0.072 & 0.135 & 0.087 \\
    Llama-3.1-70B & 0.049 & 0.080 & 0.000 & 0.192 & 0.346 & 0.133 \\
    Llama-3.3-70B & 0.061 & 0.190 & 0.000 & 0.160 & 0.105 & 0.103 \\
    Qwen3-8B & 0.456 & 0.323 & 0.311 & 0.361 & 0.484 & 0.387 \\
    Qwen3-14B & 0.544 & 0.416 & 0.237 & 0.455 & 0.559 & 0.442 \\
    Qwen3-32B & 0.629 & 0.478 & 0.286 & 0.476 & 0.589 & 0.492 \\
    \hline
    \multicolumn{7}{l}{\textit{\textbf{Closed-source}}} \\
    GPT-4o & 0.515 & 0.050 & 0.048 & 0.270 & 0.565 & 0.289 \\
    GPT-5 & 0.775 & \underline{0.618} & \textbf{0.780} & \textbf{0.698} & \textbf{0.714} & \underline{0.717} \\
    Gemini-2.5-Flash & \underline{0.788} & 0.639 & \underline{0.722} & 0.492 & 0.705 & 0.669 \\
    Gemini-2.5-Pro & \textbf{0.799} & \textbf{0.704} & 0.740 & \underline{0.687} & \underline{0.696} & \textbf{0.725} \\
    \hline
  \end{tabular}
  \caption{\label{tab:multiclass-f1} F1-scores of LLMs by hallucination type on the multi-class classification task (404 hallucination-mode samples). \textit{Overall} reflects the macro-average across five types. Bold and underlined values indicate the best and second-best scores.}
\end{table*}

Table~\ref{tab:cm-gpt5} shows confusion matrices for the two strongest base detectors (GPT-5 and Gemini-2.5-Pro) on the multi-class task. Among hallucination-type confusions, the most frequent is \textit{Contradictory}$\leftrightarrow$\textit{Unverifiable}, with a smaller but notable share of \textit{False Acceptance} misclassified as \textit{Unverifiable}.

\begin{table*}[h]
\centering
\small
\begin{tabular}{lrrrrrr}
\toprule
\textbf{GT / Pred} & \textbf{No Hallu.} & \textbf{Contradictory} & \textbf{Unverifiable} & \textbf{Irrelevance} & \textbf{False Refusal} & \textbf{False Acceptance} \\
\midrule
\multicolumn{7}{l}{\textit{\textbf{GPT-5}}} \\
No Hallucination & 360 & 7 & 3 & 3 & 29 & 2 \\
Contradictory & 24 & 110 & 20 & 4 & 0 & 0 \\
Unverifiable & 21 & 4 & 51 & 2 & 0 & 0 \\
Irrelevance & 18 & 2 & 2 & 55 & 0 & 0 \\
False Refusal & 7 & 0 & 0 & 0 & 45 & 0 \\
False Acceptance & 3 & 3 & 11 & 0 & 0 & 22 \\
\midrule
\multicolumn{7}{l}{\textit{\textbf{Gemini-2.5-Pro}}} \\
No Hallucination & 359 & 7 & 1 & 11 & 22 & 4 \\
Contradictory & 24 & 115 & 10 & 7 & 1 & 0 \\
Unverifiable & 16 & 3 & 56 & 2 & 0 & 1 \\
Irrelevance & 12 & 3 & 5 & 57 & 0 & 0 \\
False Refusal & 12 & 0 & 0 & 0 & 40 & 0 \\
False Acceptance & 5 & 2 & 9 & 0 & 0 & 23 \\
\bottomrule
\end{tabular}
\caption{Confusion matrices for GPT-5 and Gemini-2.5-Pro on the multi-class task. Rows are ground truth, columns are predictions.}
\label{tab:cm-gpt5}
\end{table*}

\subsection{Cross-source Detection Accuracy}
\label{appendix:cross-source-table}

Table~\ref{tab:cross-source} reports binary detection accuracy (recall) by hallucination type and across two injection sources, GPT-4o and Gemini-2.5-Flash. Detector rankings are largely preserved across sources, and per-type gaps remain modest ($\pm$10\%p) for most hallucination types, with the exception of \textit{Irrelevance}, which shows a pronounced difference.

\begin{table*}[t]
\centering
\small
\setlength{\tabcolsep}{3pt}
\begin{tabular}{lcccccccccccc}
\toprule
& \multicolumn{2}{c}{\textbf{Contradictory}} & \multicolumn{2}{c}{\textbf{Unverifiable}} & \multicolumn{2}{c}{\textbf{Irrelevance}} & \multicolumn{2}{c}{\textbf{False Acc.}} & \multicolumn{2}{c}{\textbf{False Ref.}} & \multicolumn{2}{c}{\textbf{Avg.}} \\
\cmidrule(lr){2-3} \cmidrule(lr){4-5} \cmidrule(lr){6-7} \cmidrule(lr){8-9} \cmidrule(lr){10-11} \cmidrule(lr){12-13}
\textbf{Model} & \textbf{GPT} & \textbf{Gem} & \textbf{GPT} & \textbf{Gem} & \textbf{GPT} & \textbf{Gem} & \textbf{GPT} & \textbf{Gem} & \textbf{GPT} & \textbf{Gem} & \textbf{GPT} & \textbf{Gem} \\
\midrule
kanana-2-30b & 0.92 & 0.92 & 0.97 & 0.95 & 0.95 & 0.95 & 1.00 & 0.89 & 0.33 & 0.39 & 0.83 & 0.82 \\
EXAONE-4.0-32B & 0.67 & 0.72 & 0.74 & 0.72 & 0.93 & 0.70 & 1.00 & 1.00 & 0.62 & 0.79 & 0.79 & 0.79 \\
Llama-3.1-8B & 0.94 & 0.97 & 0.97 & 0.92 & 0.97 & 0.89 & 1.00 & 0.95 & 1.00 & 1.00 & 0.98 & 0.95 \\
Llama-3.1-70B & 0.85 & 0.85 & 0.79 & 0.69 & 0.93 & 0.76 & 0.90 & 0.84 & 0.92 & 0.82 & 0.88 & 0.79 \\
Llama-3.3-70B & 0.87 & 0.87 & 0.79 & 0.72 & 0.93 & 0.78 & 0.95 & 0.89 & 1.00 & 0.93 & 0.91 & 0.84 \\
Qwen3-8B & 0.53 & 0.54 & 0.49 & 0.33 & 0.60 & 0.24 & 0.80 & 0.63 & 0.71 & 0.71 & 0.63 & 0.49 \\
Qwen3-14B & 0.70 & 0.68 & 0.59 & 0.31 & 0.55 & 0.38 & 0.90 & 0.63 & 0.54 & 0.46 & 0.66 & 0.49 \\
Qwen3-32B & 0.65 & 0.73 & 0.59 & 0.36 & 0.68 & 0.22 & 0.80 & 0.74 & 0.67 & 0.75 & 0.68 & 0.56 \\
GPT-4o & 0.71 & 0.72 & 0.59 & 0.64 & 1.00 & 0.76 & 1.00 & 1.00 & 0.92 & 0.89 & 0.84 & 0.80 \\
GPT-5 & 0.84 & 0.85 & 0.77 & 0.62 & 0.82 & 0.41 & 0.90 & 0.74 & 0.92 & 0.86 & 0.85 & 0.70 \\
Gem-2.5-Flash & 0.90 & 0.87 & 0.87 & 0.69 & 0.93 & 0.68 & 0.85 & 0.68 & 0.88 & 0.89 & 0.89 & 0.76 \\
Gem-2.5-Pro & 0.86 & 0.84 & 0.82 & 0.59 & 0.93 & 0.68 & 0.95 & 0.74 & 0.71 & 0.71 & 0.85 & 0.71 \\
\bottomrule
\end{tabular}
\caption{Binary detection accuracy (recall) by hallucination type and injection source (GPT = GPT-4o-injected, Gem = Gemini-injected). Each cell shows the proportion of hallucinated samples correctly detected.}
\label{tab:cross-source}
\end{table*}

\section{Hallucination Injection and Detection Prompts}
\label{appendix:prompt_templates}

This appendix provides supplementary details for the hallucination-injection pipeline described in Section~\ref{sec:injection}, along with the prompt templates used for both hallucination injection and detection. Table~\ref{tab:injection} lists the injection strategies for each hallucination type and subtype. Table~\ref{tab:pairing} maps each (Retrieval Context, Response) configuration to its corresponding type label.

\textit{False Refusal} and \textit{False Acceptance} are constructed by response and passage substitution, respectively, and do not require LLM prompting. The remaining three injection types---\textit{Contradictory}, \textit{Unverifiable}, and \textit{Irrelevance}---use prompted generation that elicits a rationale alongside the modified response. Table~\ref{tab:positioning-prompts} gives the turn-position selection prompt. Tables~\ref{tab:contradictory-basic-prompts}, \ref{tab:contradictory-inconsistency-prompts}, \ref{tab:unverifiable-basic-prompts}, and \ref{tab:irrelevance-basic-prompts} present the three injection prompts (with separate \textit{Contradictory} prompts for context contradiction and dialogue-history inconsistency). Tables~\ref{tab:hallucination-binary-prompt}, \ref{tab:hallucination-multiclass-prompt}, and \ref{tab:hallucination-fourclass-prompt} present the binary, multi-class, and four-class detection prompts.

\begin{table*}[t]
\centering
\small
\begin{tabular}{lp{3.5cm}p{9cm}}
\toprule
\textbf{Type} & \textbf{Subtype / Strategy} & \textbf{Injection Logic} \\ \midrule
\textbf{Contradictory} & Financial Term Misunderstanding & Swap a financial term with a similar but distinct one (e.g., ``Credit Loan'' $\leftrightarrow$ ``Mortgage Loan''). \\
& Modifier Change & Flip or omit critical qualifiers such as ``only,'' ``minimum,'' or ``within'' to distort legal boundaries. \\
& Number Error & Perturb figures (rates, dates, limits) using plausible values or conflicting data from the context. \\
& Inconsistency & Intentionally generate a response that contradicts statements previously made in $H_n$. \\ \midrule
\textbf{Irrelevance} & Condition Not Satisfied & Ignore query constraints regarding target demographics, periods, or eligibility requirements. \\
& Off-topic Drift & Discuss a different aspect of the same subject (e.g., answering ``period'' when asked about ``method''). \\
& Question Misunderstanding & Misinterpret grammatical particles or interrogatives (e.g., confusing ``from'' with ``until''). \\ \midrule
\textbf{Unverifiable} & Fabricated Detail & Insert non-existent procedures, exceptions, or clauses not supported by $p^*$ (under 5 words). \\
& Subjective Opinion & Interject unfounded evaluations, recommendations, or judgments into otherwise neutral facts. \\ \midrule
\textbf{False Acceptance} & - & Respond to a \textit{hard negative} $p^-$ as if it contains valid evidence. \\
\midrule
\textbf{False Refusal} & - & Replace a fully supported response with a fixed refusal string despite the presence of evidence. \\ \bottomrule
\end{tabular}
\caption{Injection strategies for each hallucination type and subtype in K-FinHallu.}
\label{tab:injection}
\end{table*}

\begin{table*}[t]
\centering
\small
\setlength{\tabcolsep}{8pt}
\begin{tabular}{lll}
\toprule
\textbf{Retrieval Context ($\mathcal{P}_n$)} & \textbf{Response ($a_n$)} & \textbf{Type} \\
\midrule
Contains Positive Passages & Original & Faithful Answer \\
Contains Positive Passages & Perturbed/Injected & Contradictory / Unverifiable / Irrelevance \\
Consists Entirely of Hard Negatives & Original & False Acceptance \\
Contains Positive Passages & Refusal String & False Refusal \\
Consists Entirely of Hard Negatives & Refusal String & True Refusal \\
\bottomrule
\end{tabular}
\caption{The combinatorial mapping of Retrieval Context ($\mathcal{P}_n$) and Response ($a_n$) configurations in K-FinHallu. \textit{Positive} and \textit{Negative} refer to the answerable and unanswerable $\mathcal{P}_n$ configurations defined in Section~\ref{subsec:taskformulation}.}
\label{tab:pairing}
\end{table*}


\begin{table*}[ht]
\centering
\small
\begin{tabularx}{\textwidth}{l X}
\toprule
\textbf{Type} & \textbf{Prompt} \\
\midrule
\textbf{System} & 
\textbf{[EN]} You are an expert in determining where to inject hallucinations into a dialogue. Determine the appropriate positions so that each hallucination type is naturally integrated. \\
& \\
& \textbf{[KO]} 당신은 대화에 환각을 주입할 위치를 결정하는 전문가입니다. 각 환각 유형이 적절한 위치에 주입되도록 판단하세요. \\
\midrule
\textbf{User} & 
\textbf{[EN]} The following is the full dialogue: \texttt{\{dialogue\_text\}}. \\
& Hallucination types to inject (in order): \texttt{\{hallucination\_types\}}. \\
& \texttt{\{hallucination\_definitions\}}. \\
& \\
& Decide which turn to inject each hallucination type in order. Each hallucination type must be injected in a different turn. The order of turn numbers matters (first turn number maps to type 1, second to type 2, etc.). \texttt{\{constraint\_note\}}. \\
& \\
& Respond in JSON format: \texttt{\{"turn\_indices": [1, 2, ...]\}}. \\
& turn\_indices: List of turn numbers to inject each hallucination type (in order, starting from 1). \\
\cmidrule{2-2}
& \textbf{[KO]} 다음은 전체 대화입니다: \texttt{\{dialogue\_text\}}. \\
& 주입할 환각 유형 (순서대로): \texttt{\{hallucination\_types\}}. \\
& \texttt{\{hallucination\_definitions\}}. \\
& \\
& 각 환각 유형을 순서대로 어떤 턴에 주입할지 결정하세요. 각 환각 유형은 서로 다른 턴에 주입되어야 합니다. 반환하는 턴 번호의 순서가 중요합니다 (첫 번째 턴 번호는 1번 유형에, 두 번째 턴 번호는 2번 유형에 매핑됩니다). \texttt{\{constraint\_note\}}. \\
& \\
& JSON 형식으로 답변하세요: \texttt{\{"turn\_indices": [1, 2, ...]\}}. \\
& turn\_indices: 각 환각 유형을 주입할 턴 번호 리스트 (순서대로, 1부터 시작). \\
\bottomrule
\end{tabularx}
\caption{Positioning prompts for determining hallucination injection locations.}
\label{tab:positioning-prompts}
\end{table*}

\begin{table*}[ht]
\centering
\small
\begin{tabularx}{\textwidth}{l X}
\toprule
\textbf{Type} & \textbf{Prompt} \\
\midrule
\textbf{System} & \textbf{[EN]} You are an expert at subtly modifying answers to create conflicts with context. Generate natural hallucinations with minimal modifications. \\
& \\
& \textbf{[KO]} 답변을 미묘하게 수정하여 컨텍스트와 충돌을 만드는 전문가입니다. 최소 수정으로 자연스러운 환각을 생성하세요. \\
\midrule
\textbf{User} & \textbf{[EN]} The following question and answer are given. \\
& Question: \texttt{\{user\_question\}}. \\
& Answer: \texttt{\{original\_answer\}}. \\
& Related Document: \texttt{\{passage\_text\}}. \\
& \\
& Modify the answer minimally to create a subtle contradiction with the related document/context. Change only 1-2 key financial terms/words/modifiers/numbers to distort the meaning. Do not add unverifiable information (sentences or phrases); maintain the original answer structure and length. Make the modification so subtle that it's difficult to notice, ensuring the answer appears natural. Do not change the subject of the answer as much as possible. \\
& \\
& You may refer to the following methods: \\
& - Financial term misunderstanding: Replace difficult financial terms with similar but different terms to distort meaning (e.g., ``unsecured loan'' $\leftrightarrow$ ``secured loan'', ``equal installment'' $\rightarrow$ ``equal principal'', ``interest rate'' $\rightarrow$ ``yield''). \\
& - Modifier change or omission: Change/omit modifiers like only/just/minimum/maximum/at least/at most/over/under/partially. \\
& - Number error: Replace key numbers (period/rate/limit/frequency) with similar range numbers, or replace with other related numbers from context. \\
& \\
& Respond in the following format (first explain the rationale, then write the final answer): \\
& Rationale: [Explain which part was modified and why]. \\
& Answer: [Modified answer]. \\
\cmidrule{2-2}
& \textbf{[KO]} 다음 질문과 답변이 주어집니다. \\
& 질문: \texttt{\{user\_question\}}. \\
& 답변: \texttt{\{original\_answer\}}. \\
& 관련 문서: \texttt{\{passage\_text\}}. \\
& \\
& 답변을 최소한으로만 수정하여, 관련 문서/컨텍스트와 미묘하게 충돌(모순)되도록 환각을 만드세요. 전체 답변을 다시 작성하지 말고, 핵심 금융전문용어/단어/수식어/숫자 중 1~2개만 바꿔서 의미가 틀어지게 하세요. 검증 불가능한 정보(문장이나 구)를 추가하지말고, 기존 답변 구조와 길이를 유지하세요. 알아차리기 어려울 정도로 미묘하게 수정하여, 답변이 자연스럽게 보이도록 하세요. 답변의 주어는 최대한 바꾸지마세요. \\
& \\
& 다음과 같은 방법을 참고할 수 있습니다: \\
& - 금융 전문 용어 오해: 답변에 있는 어려운 금융 전문 용어를 유사하지만 다른 금융 전문 용어로 바꿔 의미를 왜곡 (예: ``신용대출''$\leftrightarrow$``담보대출'', ``원리금상환''$\rightarrow$``원금균등상환'', ``이자율''$\rightarrow$``금리''). \\
& - 한정어/수식어 변경하거나 누락: 오직/만/최소/최대/이상/이하/초과/미만/부분적으로 등 한정어/수식어를 변경/누락. \\
& - 숫자 오류: 기간/금리/한도/횟수 등 핵심 숫자를 비슷한 범위의 다른 숫자로 교체하거나, 컨텍스트 내 다른 관련 숫자로 교체. \\
& \\
& 다음 형식으로 답변하세요 (먼저 근거를 설명하고, 그 다음 최종 답변을 작성하세요): \\
& 근거: [어떤 부분을 왜 수정했는지 설명]. \\
& 답변: [수정된 답변]. \\
\bottomrule
\end{tabularx}
\caption{Contradictory prompts: Basic context contradiction type.}
\label{tab:contradictory-basic-prompts}
\end{table*}

\begin{table*}[ht]
\centering
\small
\begin{tabularx}{\textwidth}{l X}
\toprule
\textbf{Type} & \textbf{Prompt} \\
\midrule
\textbf{System} & \textbf{[EN]} You are an expert at subtly modifying answers to create conflicts with context. Generate natural hallucinations with minimal modifications. \\
& \\
& \textbf{[KO]} 답변을 미묘하게 수정하여 컨텍스트와 충돌을 만드는 전문가입니다. 최소 수정으로 자연스러운 환각을 생성하세요. \\
\midrule
\textbf{User} & \textbf{[EN]} The following dialogue history and current answer are given. \\
& Dialogue History: \texttt{\{history\_text\}}. \\
& Current Question: \texttt{\{user\_question\}}. \\
& Current Answer: \texttt{\{original\_answer\}}. \\
& Related Document: \texttt{\{passage\_text\}}. \\
& \\
& Modify the current answer minimally so that it contradicts the prior dialogue in a subtle and hard-to-notice way. Change only 1--2 key financial terms/words/modifiers/numbers to distort the meaning. Do not add unverifiable information (sentences or phrases); maintain the original answer structure and length. Make the modification so subtle that it appears natural. Do not change the subject of the answer, and do not produce a response unrelated to the question. \\
& \\
& You may refer to the following methods: \\
& - Financial term misunderstanding: Replace difficult financial terms with similar but different terms to distort meaning (e.g., ``unsecured loan'' $\leftrightarrow$ ``secured loan'', ``equal installment'' $\rightarrow$ ``equal principal'', ``interest rate'' $\rightarrow$ ``yield''). \\
& - Modifier change or omission: Change/omit modifiers like only/just/minimum/maximum/at least/at most/over/under/partially. \\
& - Number error: Replace key numbers (period/rate/limit/frequency) with similar range numbers, or replace with other related numbers from context. \\
& \\
& Respond in the following format (first explain the rationale, then write the final answer): \\
& Rationale: [Explain which part was modified and why]. \\
& Answer: [Modified answer]. \\
\cmidrule{2-2}
& \textbf{[KO]} 다음 대화 히스토리와 현재 답변이 주어집니다. \\
& 대화 히스토리: \texttt{\{history\_text\}}. \\
& 현재 질문: \texttt{\{user\_question\}}. \\
& 현재 답변: \texttt{\{original\_answer\}}. \\
& 관련 문서: \texttt{\{passage\_text\}}. \\
& \\
& 현재 답변을 이전 대화와 모순되도록 미묘하고 알아차리기 쉽지않은 환각을 만들도록 최소한으로 수정하세요. 전체 답변을 다시 작성하지 말고, 핵심 금융전문용어/단어/수식어/숫자 중 1~2개만 바꿔서 의미가 틀어지게 하세요. 검증 불가능한 정보(문장이나 구)를 추가하지말고, 기존 답변 구조와 길이를 유지하세요. 알아차리기 어려울 정도로 미묘하게 수정하여, 답변이 자연스럽게 보이도록 하세요. 답변의 주어는 최대한 바꾸지마세요. 미묘하게 수정한다고 해서 질문과 관련 없는 대답을 하면 안 됩니다. \\
& \\
& 다음과 같은 방법을 참고할 수 있습니다: \\
& - 금융 전문 용어 오해: 답변에 있는 어려운 금융 전문 용어를 유사하지만 다른 금융 전문 용어로 바꿔 의미를 왜곡 (예: ``신용대출''$\leftrightarrow$``담보대출'', ``원리금상환''$\rightarrow$``원금균등상환'', ``이자율''$\rightarrow$``금리''). \\
& - 한정어/수식어 변경하거나 누락: 오직/만/최소/최대/이상/이하/초과/미만/부분적으로 등 한정어/수식어를 변경/누락. \\
& - 숫자 오류: 기간/금리/한도/횟수 등 핵심 숫자를 비슷한 범위의 다른 숫자로 교체하거나, 컨텍스트 내 다른 관련 숫자로 교체. \\
& \\
& 다음 형식으로 답변하세요 (먼저 근거를 설명하고, 그 다음 최종 답변을 작성하세요): \\
& 근거: [어떤 부분을 왜 수정했는지 설명]. \\
& 답변: [수정된 답변]. \\
\bottomrule
\end{tabularx}
\caption{Contradictory prompts: Inconsistency with prior dialogue history.}
\label{tab:contradictory-inconsistency-prompts}
\end{table*}

\begin{table*}[ht]
\centering
\small
\begin{tabularx}{\textwidth}{l X}
\toprule
\textbf{Type} & \textbf{Prompt} \\
\midrule
\textbf{System} & \textbf{[EN]} You are an expert at naturally adding unverifiable information. Insert only short phrases of 3-5 words. \\
& \\
& \textbf{[KO]} 근거 없는 정보를 자연스럽게 추가하는 전문가입니다. 3-5어절의 짧은 phrase만 삽입하세요. \\
\midrule
\textbf{User} & \textbf{[EN]} The following question and answer are given. \\
& Question: \texttt{\{user\_question\}}. \\
& Original Answer: \texttt{\{original\_answer\}}. \\
& Related Document: \texttt{\{passage\_text\}}. \\
& \\
& The original answer is correct. Add information that cannot be verified by the document/context as a subtle, hard-to-detect hallucination. Do not add a whole sentence; only add short content of less than 5 words. Maintain the structure and content of the original answer as much as possible. Write the added information so subtly that it appears real and is difficult to notice. \\
& \\
& You may refer to the following methods: \\
& - Unverifiable detailed information: Insert short details like procedures/conditions/exceptions/periods that are not in the document. \\
& - Unverifiable subjective opinion: Insert evaluations/recommendations that are not in the document to appear factual. \\
& \\
& Respond in the following format (first explain the rationale, then write the final answer): \\
& Rationale: [Explain which part was modified and why]. \\
& Answer: [Modified answer]. \\
\cmidrule{2-2}
& \textbf{[KO]} 다음 질문과 답변이 주어집니다. \\
& 질문: \texttt{\{user\_question\}}. \\
& 원본 답변: \texttt{\{original\_answer\}}. \\
& 관련 문서: \texttt{\{passage\_text\}}. \\
& \\
& 원본 답변은 올바릅니다. 이 답변에 문서/컨텍스트로 검증 불가한 정보를 미묘하고 알아차리기 쉽지않게 추가하세요. 한 문장 전체를 추가하지 말고, 5어절 미만의 짧은 내용을 추가하세요. 원본 답변의 구조와 내용은 최대한 유지하세요. 추가된 정보가 실제 정보처럼 보이도록 알아차리기 어려울 정도로 미묘하게 작성하세요. \\
& \\
& 다음과 같은 방법을 참고할 수 있습니다: \\
& - 근거 없는 상세 정보: 문서에 없는 절차/조건/예외/기간 등 짧은 디테일을 끼워 넣기. \\
& - 근거 없는 주관적 의견: 문서에 없는 평가/추천을 사실처럼 보이게 끼워 넣기. \\
& \\
& 다음 형식으로 답변하세요 (먼저 근거를 설명하고, 그 다음 최종 답변을 작성하세요): \\
& 근거: [어떤 부분을 왜 수정했는지 설명]. \\
& 답변: [수정된 답변]. \\
\bottomrule
\end{tabularx}
\caption{Unverifiable prompts: Basic unverifiable information addition type.}
\label{tab:unverifiable-basic-prompts}
\end{table*}

\begin{table*}[ht]
\centering
\small
\begin{tabularx}{\textwidth}{l X}
\toprule
\textbf{Type} & \textbf{Prompt} \\
\midrule
\textbf{System} & \textbf{[EN]} You are an expert at creating answers that deviate from the question intent. Maintain the core topic but answer something unrelated to the question. \\
& \\
& \textbf{[KO]} 질문 의도와 어긋나는 답변을 만드는 전문가입니다. 핵심 토픽은 유지하되 질문과 관련 없는 답을 하세요. \\
\midrule
\textbf{User} & \textbf{[EN]} The following conversation history, current question, and answer are given. \\
& Conversation History: \texttt{\{history\_text\}}. \\
& Current Question: \texttt{\{user\_question\}}. \\
& Current Answer: \texttt{\{original\_answer\}}. \\
& Related Document: \texttt{\{passage\_text\}}. \\
& \\
& Modify the answer minimally to deviate from the question intent while maintaining key keywords within the same topic. \\
& You may refer to the following methods: \\
& - Condition mismatch: Modify to not satisfy the question's conditions (target/period/scope/qualification) (e.g., change ``3 months'' to ``1 year'' when asked about ``3 months''). \\
& - Topic drift: Answer about a different aspect within the same topic (e.g., answer about ``application period'' when asked about ``application method''). \\
& - Question misunderstanding: Misinterpret grammatical elements (particles/interrogatives) and answer differently (e.g., misunderstand ``from'' as ``until'' and answer about ``until when''). \\
& \\
& Respond in the following format (first explain the rationale, then write the final answer): \\
& Rationale: [Explain which method was used, which part was modified, and why]. \\
& Answer: [Modified answer]. \\
\cmidrule{2-2}
& \textbf{[KO]} 다음 대화 히스토리와 현재 질문, 답변이 주어집니다. \\
& 대화 히스토리: \texttt{\{history\_text\}}. \\
& 현재 질문: \texttt{\{user\_question\}}. \\
& 현재 답변: \texttt{\{original\_answer\}}. \\
& 관련 문서: \texttt{\{passage\_text\}}. \\
& \\
& 같은 주제 내에서 질문의 핵심 키워드를 유지하면서 질문 의도와 어긋나도록 답변을 최소한으로만 수정하세요. \\
& 다음과 같은 방법을 참고할 수 있습니다: \\
& - 조건 불만족: 질문의 조건(대상/기간/범위/자격)을 만족하지 않도록 수정 (예: ``3개월''을 묻는데 ``1년''으로 답하도록 변경). \\
& - 주제 이탈: 같은 주제 내에서 질문과 다른 측면에 대해 답변 (예: ``신청 방법''을 묻는데 ``신청 기간''에 대해 답변). \\
& - 질문 오해: 문법적 요소(조사/의문사)를 오해하여 다르게 답변 (예: ``부터''를 ``까지''로 오해하여 ``언제까지''에 대해 답변). \\
& \\
& 다음 형식으로 답변하세요 (먼저 근거를 설명하고, 그 다음 최종 답변을 작성하세요): \\
& 근거: [어떤 방법을 사용하여 어떤 부분을 왜 수정했는지 설명]. \\
& 답변: [수정된 답변]. \\
\bottomrule
\end{tabularx}
\caption{Irrelevance prompts: Basic question intent deviation type.}
\label{tab:irrelevance-basic-prompts}
\end{table*}

\begin{table*}[ht]
\centering
\small
\begin{tabularx}{\textwidth}{X}
\toprule
\textbf{Prompt} \\
\midrule
\textbf{[EN]} \\
\textbf{Task:} \\
Determine whether the current response in the following dialogue is a hallucination. \\
\textbf{Important:} Judge based only on the given dialogue history and related documents. Do not judge using the LLM's internal knowledge. \\
\\
\textbf{Classification Criteria:} \\
It is a hallucination if any of the following applies: \\
- When the answer directly contradicts the given documents or previous dialogue history \\
- When the answer includes information that cannot be verified from the given documents \\
- When the response is not logically connected to the user's intent or question \\
- When the answer refuses to respond despite sufficient evidence in the given documents \\
- When an answer is provided despite the situation being unanswerable from the given documents \\
\\
It is NOT a hallucination only in the following cases: \\
- When the answer correctly responds to the question based on information from the given documents \\
- When the answer refuses to respond, acknowledging insufficient evidence in the given documents \\
\\
\textbf{Output Format:} \\
Output only 0 or 1, without any other explanation. \\
\\
0: Not Hallucination \\
1: Hallucination \\
\\
\textbf{Dialogue:} \texttt{\{dialogue\}} \\
\textbf{Question:} \texttt{\{question\}} \\
\textbf{Related Documents:} \texttt{\{contexts\}} \\
\textbf{Answer:} \texttt{\{answer\}} \\
\midrule
\textbf{[KO]} \\
\textbf{태스크:} \\
다음에 제시되는 대화에서 현재 응답이 할루시네이션인지 판별하세요. \\
\textbf{중요:} 주어진 대화 히스토리와 관련 문서만을 기준으로 판단하세요. LLM의 내부 지식으로 판단하지 마세요. \\
\\
\textbf{분류 기준:} \\
다음 중 하나라도 해당하면 할루시네이션입니다: \\
- 답변이 주어진 문서나 이전 대화 히스토리와 직접적으로 모순되는 경우 \\
- 답변에 주어진 문서에 검증 불가능한 정보를 포함하는 경우 \\
- 사용자의 의도나 질문과 논리적으로 연결되지 않는 응답을 하는 경우 \\
- 주어진 문서에 충분한 증거가 있음에도 불구하고 답변을 거부하는 경우 \\
- 주어진 문서로 답변할 수 없는 상황임에도 불구하고 답변을 제공하는 경우 \\
\\
다음의 경우만 할루시네이션이 아닙니다: \\
- 주어진 문서의 정보를 바탕으로 질문에 올바르게 답변한 경우 \\
- 주어진 문서에 증거가 부족함을 인정하고 답변을 거부하는 경우 \\
\\
\textbf{출력 형식:} \\
다른 설명 없이 오직 0 또는 1만 출력하세요. \\
\\
0: 할루시네이션 아님 \\
1: 할루시네이션 \\
\\
\textbf{대화:} \texttt{\{dialogue\}} \\
\textbf{질문:} \texttt{\{question\}} \\
\textbf{관련 문서:} \texttt{\{contexts\}} \\
\textbf{답변:} \texttt{\{answer\}} \\
\bottomrule
\end{tabularx}
\caption{Binary hallucination detection prompt template.}
\label{tab:hallucination-binary-prompt}
\end{table*}

\begin{table*}[ht]
\centering
\small
\begin{tabularx}{\textwidth}{X}
\toprule
\textbf{Prompt} \\
\midrule
\textbf{[EN]} \\
\textbf{Task:} \\
Determine whether the current response in the following dialogue is a hallucination, and if so, classify its type. \\
\textbf{Important:} Judge based only on the given dialogue history and related documents. Do not judge using the LLM's internal knowledge. \\
\\
\textbf{Classification Criteria:} \\
When the given documents contain sufficient evidence for the question: \\
- No Hallucination: When the answer correctly responds to the question based on information from the given documents \\
- Contradictory: When the answer directly contradicts the given documents or previous dialogue history \\
- Unverifiable: When the answer includes information that cannot be verified from the given documents \\
- Irrelevance: When the response is not logically connected to the user's intent or question \\
- False Refusal: When the answer refuses to respond despite sufficient evidence in the given documents \\
\\
When the given documents lack sufficient evidence for the question: \\
- False Acceptance: When an answer is provided despite the situation being unanswerable from the given documents \\
- No Hallucination: When the answer refuses to respond, acknowledging insufficient evidence in the given documents \\
\\
\textbf{Output Format:} \\
Output only a single number between 0 and 5 representing the hallucination type, without any other explanation. \\
\\
0: No Hallucination \\
1: Contradictory \\
2: Unverifiable \\
3: Irrelevance \\
4: False Refusal \\
5: False Acceptance \\
\\
\textbf{Dialogue:} \texttt{\{dialogue\}} \\
\textbf{Question:} \texttt{\{question\}} \\
\textbf{Related Documents:} \texttt{\{contexts\}} \\
\textbf{Answer:} \texttt{\{answer\}} \\
\midrule
\textbf{[KO]} \\
\textbf{태스크:} \\
다음에 제시되는 대화에서 현재 응답이 할루시네이션인지 판별하고, 할루시네이션인 경우 유형을 분류하세요. \\
\textbf{중요:} 주어진 대화 히스토리와 관련 문서만을 기준으로 판단하세요. LLM의 내부 지식으로 판단하지 마세요. \\
\\
\textbf{분류 기준:} \\
주어진 문서에 질문에 대한 충분한 증거가 있는 경우: \\
- No Hallucination: 주어진 문서의 정보를 바탕으로 질문에 올바르게 답변한 경우 \\
- Contradictory: 답변이 주어진 문서나 이전 대화 히스토리와 직접적으로 모순되는 경우 \\
- Unverifiable: 답변에 주어진 문서에 검증 불가능한 정보를 포함하는 경우 \\
- Irrelevance: 사용자의 의도나 질문과 논리적으로 연결되지 않는 응답을 하는 경우 \\
- False Refusal: 주어진 문서에 충분한 증거가 있음에도 불구하고 답변을 거부하는 경우 \\
\\
주어진 문서에 질문에 대한 충분한 증거가 없는 경우: \\
- False Acceptance: 주어진 문서로 답변할 수 없는 상황임에도 불구하고 답변을 제공하는 경우 \\
- No Hallucination: 주어진 문서에 증거가 부족함을 인정하고 답변을 거부하는 경우 \\
\\
\textbf{출력 형식:} \\
다른 설명 없이 오직 할루시네이션 유형을 나타내는 0에서 5 사이의 숫자 하나만 출력하면 됩니다. \\
\\
0: No Hallucination \\
1: Contradictory \\
2: Unverifiable \\
3: Irrelevance \\
4: False Refusal \\
5: False Acceptance \\
\\
\textbf{대화:} \texttt{\{dialogue\}} \\
\textbf{질문:} \texttt{\{question\}} \\
\textbf{관련 문서:} \texttt{\{contexts\}} \\
\textbf{답변:} \texttt{\{answer\}} \\
\bottomrule
\end{tabularx}
\caption{Multi-class hallucination detection and classification prompt template.}
\label{tab:hallucination-multiclass-prompt}
\end{table*}

\begin{table*}[ht]
\centering
\small
\begin{tabularx}{\textwidth}{X}
\toprule
\textbf{Prompt} \\
\midrule
\textbf{[EN]} \\
\textbf{Task:} \\
Classify the current response in the following conversation into one of four types. \\
\textbf{Important:} Judge only based on the provided dialogue history and reference documents. Do not use the LLM's internal knowledge. \\
\\
\textbf{Classification Criteria:} \\
When the reference documents contain sufficient evidence for the question: \\
- Faithful Answer: The response correctly answers the question based on the information in the reference documents. \\
- Hallucination: Any of the following applies: \\
\quad{}* The response directly contradicts the reference documents or prior dialogue history. \\
\quad{}* The response contains information that cannot be verified from the reference documents. \\
\quad{}* The response is logically irrelevant to the user's intent or question. \\
- False Refusal: The response refuses to answer despite sufficient evidence being available in the reference documents. \\
\\
When the reference documents do NOT contain sufficient evidence for the question: \\
- Hallucination: The response provides an answer even though the reference documents cannot support it. \\
- True Refusal: The response appropriately refuses to answer, acknowledging the lack of evidence in the reference documents. \\
\\
\textbf{Output Format:} \\
Output only a single number between 0 and 3 with no other explanation. \\
\\
0: Faithful Answer \\
1: Hallucination \\
2: False Refusal \\
3: True Refusal \\
\\
\textbf{Dialogue:} \texttt{\{dialogue\}} \\
\textbf{Question:} \texttt{\{question\}} \\
\textbf{Reference Documents:} \texttt{\{contexts\}} \\
\textbf{Answer:} \texttt{\{answer\}} \\
\midrule
\textbf{[KO]} \\
\textbf{태스크:} \\
다음에 제시되는 대화에서 현재 응답을 4가지 유형으로 분류하세요. \\
\textbf{중요:} 주어진 대화 히스토리와 관련 문서만을 기준으로 판단하세요. LLM의 내부 지식으로 판단하지 마세요. \\
\\
\textbf{분류 기준:} \\
주어진 문서에 질문에 대한 충분한 증거가 있는 경우: \\
- Faithful Answer: 주어진 문서의 정보를 바탕으로 질문에 올바르게 답변한 경우 \\
- Hallucination: 다음 중 하나라도 해당하는 경우 \\
\quad{}* 답변이 주어진 문서나 이전 대화 히스토리와 직접적으로 모순되는 경우 \\
\quad{}* 답변에 주어진 문서에 검증 불가능한 정보를 포함하는 경우 \\
\quad{}* 사용자의 의도나 질문과 논리적으로 연결되지 않는 응답을 하는 경우 \\
- False Refusal: 주어진 문서에 충분한 증거가 있음에도 불구하고 답변을 거부하는 경우 \\
\\
주어진 문서에 질문에 대한 충분한 증거가 없는 경우: \\
- Hallucination: 주어진 문서로 답변할 수 없는 상황임에도 불구하고 답변을 제공하는 경우 \\
- True Refusal: 주어진 문서에 증거가 부족함을 인정하고 적절하게 답변을 거부하는 경우 \\
\\
\textbf{출력 형식:} \\
다른 설명 없이 오직 0에서 3 사이의 숫자 하나만 출력하세요. \\
\\
0: Faithful Answer \\
1: Hallucination \\
2: False Refusal \\
3: True Refusal \\
\\
\textbf{대화:} \texttt{\{dialogue\}} \\
\textbf{질문:} \texttt{\{question\}} \\
\textbf{관련 문서:} \texttt{\{contexts\}} \\
\textbf{답변:} \texttt{\{answer\}} \\
\bottomrule
\end{tabularx}
\caption{Four-class hallucination detection prompt template.}
\label{tab:hallucination-fourclass-prompt}
\end{table*}

\end{document}